\documentclass{article}

\usepackage[preprint]{neurips_2026}

\usepackage[utf8]{inputenc} 
\usepackage[T1]{fontenc}    
\usepackage{hyperref}       
\usepackage{url}            
\usepackage{booktabs}       
\usepackage{amsfonts}       
\usepackage{nicefrac}       
\usepackage{microtype}      
\usepackage{xcolor}         

\usepackage{graphicx}
\usepackage{amsmath}
\usepackage{amssymb}
\usepackage{subcaption}
\usepackage{multirow}
\usepackage{tabularx}
\usepackage{pifont}
\usepackage{tikz}
\usetikzlibrary{calc}
\usepackage{pgfplots}
\usepgfplotslibrary{statistics}
\pgfplotsset{compat=1.18}
\usepgfplotslibrary{fillbetween}
\usepackage[ruled,linesnumbered]{algorithm2e}
\usepackage[capitalise,noabbrev]{cleveref}

\newcolumntype{C}{>{\centering\arraybackslash}X}
\newcolumntype{Y}{>{\centering\arraybackslash}X}

\DeclareMathOperator*{\argmax}{arg\,max}
\DeclareMathOperator*{\argtopk}{arg\,topk}

\usepackage{xspace}
\usepackage{soul}
\newcommand{\hlc}[2][yellow]{\sethlcolor{#1}\hl{#2}}
\newcommand{\eg}{\textit{e.g.}\xspace}
\newcommand{\ie}{\textit{i.e.}\xspace}

\newcommand{\etal}{\textit{et al.}\xspace}

\definecolor{alphaColor}{RGB}{228,26,28}
\definecolor{betaColor}{RGB}{55,126,184}
\definecolor{gammaColor}{RGB}{77,175,74}
\newcommand{\instalpha}{{\color{alphaColor}{\alpha}}}
\newcommand{\instbeta}{{\color{betaColor}{\beta}}}

\title{Rethinking Test-Time Scaling for\\Generative Flow Matching Models}

\author{
\textbf{Qingtao Yu}$^{\instalpha}$ \quad \textbf{Changlin Song}$^{\instalpha}$ \quad \textbf{Minghao Sun}$^{\instalpha}$ \quad \textbf{Zhengyang Yu}$^{\instalpha}$ \quad \textbf{Vinay Kumar Verma}$^{\instbeta}$ \\
\textbf{Soumya Roy}$^{\instbeta}$ \quad \textbf{Sumit Negi}$^{\instbeta}$ \quad \textbf{Hongdong Li}$^{\instalpha, \instbeta}$ \quad \textbf{Dylan Campbell}$^{\instalpha}$ \\[0.5em]
$^{\instalpha}$ Australian National University \quad $^{\instbeta}$ Amazon Research
}

\begin{document}

\maketitle

\vspace{-1cm}
\begin{figure}[h!]
    \centering
    \makebox[0.02\linewidth][c]{\raisebox{1em}{\rotatebox{90}{\small \; \textbf{Ours} \quad \quad \; Best others \quad \quad No TTS}}}\hfill
    \includegraphics[width=0.135\linewidth]{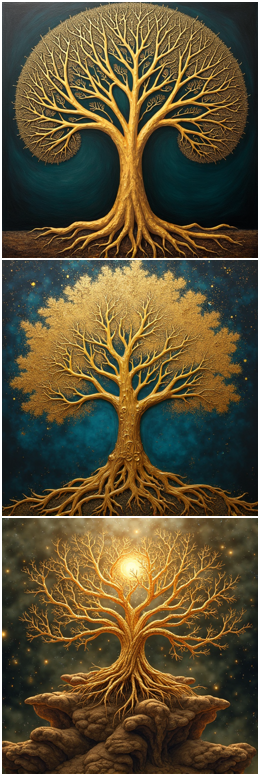}\hfill
    \includegraphics[width=0.135\linewidth]{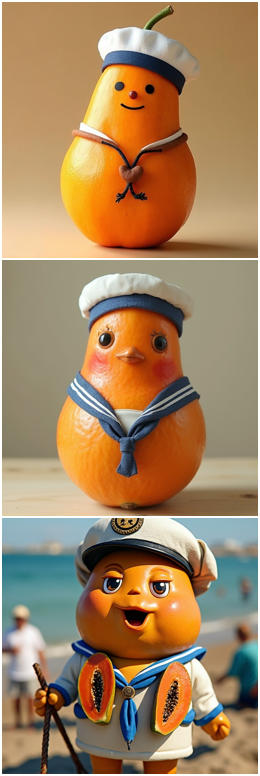}\hfill
    \includegraphics[width=0.135\linewidth]{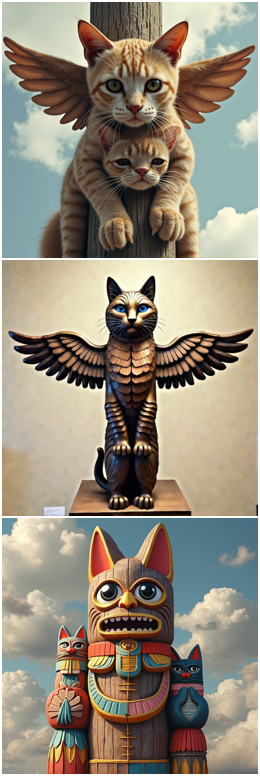}\hfill
    \includegraphics[width=0.135\linewidth]{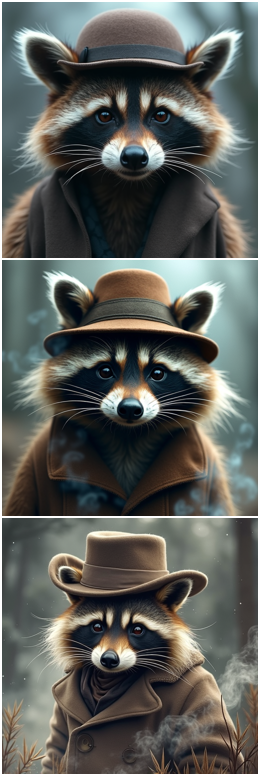}\hfill
    \includegraphics[width=0.135\linewidth]{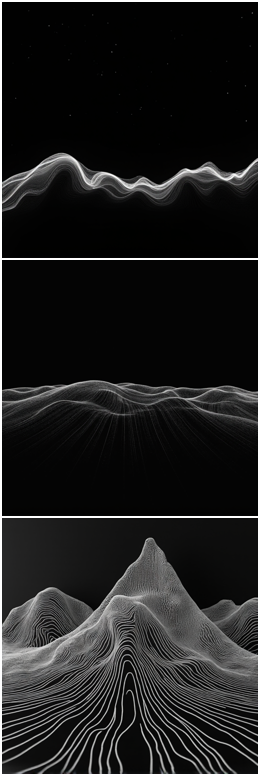}\hfill
    \includegraphics[width=0.135\linewidth]{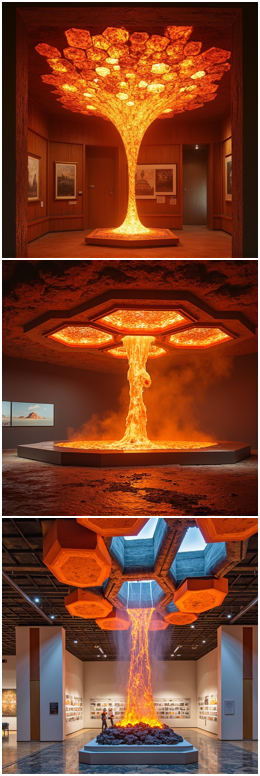}\hfill
    \includegraphics[width=0.135\linewidth]{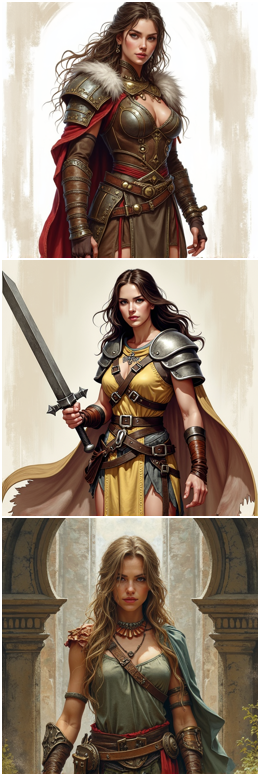}
    \caption{%
        Some qualitative comparisons.
        Top row: Flux.1-dev without test-time scaling. 
        Second and bottom row: the best outcome of other competing test-time scaling methods, and ours.
        Prompts from left to right:
        \textit{
        \hlc[pink!30]{1.} The image depicts a gold filigree tree of life in a detailed fantasy painting;
        \hlc[yellow!40]{2.} A papaya fruit dressed as a sailor;
        \hlc[green!20]{3.} Totem pole made out of cats;
        \hlc[cyan!20]{4.} The image is of a raccoon wearing a Peaky Blinders hat, surrounded by swirling mist and rendered with fine detail;
        \hlc[pink!30]{5.} White lines depict topography on a black background;
        \hlc[yellow!40]{6.} Molten lava hanging from the ceiling creates art with octagons in a museum setting;
        \hlc[green!20]{7.} A female human barbarian depicted in a traditional Dungeons and Dragons illustration.
        }
    }
    \label{fig:teaser}
\end{figure}
\vspace{-0.1cm}

\begin{abstract}
\vspace{-0.2cm}
The performance of text-to-image diffusion models may be improved at test-time by scaling computation to search for a generated image that maximizes a given reward function.
While existing trajectory-level exploration methods improve the effectiveness of test-time scaling for standard diffusion models, they are largely incompatible with modern flow matching models, which use deterministic sampling.
This imposes significant computational overhead on local trajectory search, making the trade-offs less favorable compared to global search.
However, global search strategies like trajectory pruning face two critical challenges: the sharp, low-diversity distributions characteristic of scaled flow models that restrict the candidate search space, and the bias of reward models in the early denoising process.
To overcome these limitations, we propose
Repel, a token-level mechanism that encourages sample diversity, and
NARF, a noise-aware reward fine-tuning strategy to obtain more accurate reward ranking at early denoising stages.
Together, these promote more effective test-time scaling resource allocation. 
Overall, we name our pipeline as \textbf{DOG-Trim}: \textbf{D}iversity enhanced \textbf{O}rder aligned \textbf{G}lobal flow Trimming.
The experiments demonstrate that, under the same compute cost, our approach achieves around
twice the performance improvement relative to the scaling-free baseline compared to the best existing method. 
Github: https://github.com/TerrysLearning/DOGTrimTTS.
\end{abstract}

\section{Introduction}
\label{sec:intro}
The success of large generative models is closely tied to scaling model capacities and data size during training~\cite{scalinglaw, qwen, gpt4, gemini, sd3, qwenimage, veo3}.
An orthogonal direction for improving model performance has gained increasing attention: scaling at test time~\cite{simples1, ttscode, expand, ttMeta}.
Test-time scaling (TTS) seeks to improve the output quality of a pretrained model by allocating additional computation during inference, without modifying the model itself.
While TTS has been widely explored for large language models, such as iterative reasoning~\cite{cot, cot2, cot3} or repeated sampling~\cite{snellscale, moneyllmtts, mixagent}, recent work has begun to extend this paradigm to diffusion and flow matching models for text-to-image generation~\cite{tts_ma, sana1_5, tts_var, tts_tut}.
In this setting, TTS is naturally formulated as a search problem, exploiting the stochasticity inherent in the generative process to find a sample that maximizes a given reward function.

The primary objective when designing TTS methods is to maximize generation quality under a constrained computational budget.
The simplest baseline, best-of-$N$ \cite{tts_ma, sana1_5}, generates $N$ independent samples from different initial noise latents and selects the highest-scoring outcome.
However, this approach is highly inefficient; each candidate must be fully denoised before its quality can be verified, which restricts the number of candidates that can be explored under a fixed budget.
Existing methods improve upon this by introducing local-search and trajectory-level exploration strategies~\cite{tts_greedy, dynamic_search, fk, classic, tts_tree} that concentrate search effort in promising regions of the generation space and avoid full-denoising computations for each candidate.

However, most of these strategies are designed for stochastic diffusion models and are largely incompatible with modern flow matching models~\cite{flux, sd3, sana}.
Although both paradigms denoise from a Gaussian prior, flow models generally rely on deterministic ODE solvers, which introduce benefits such as lower timestep sampling and efficient training \cite{flow_fast, flow_gen}.
However, this removes the natural mechanism for local trajectory branching and perturbation.
The alternatives, such as replacing the ODE sampler with an SDE \cite{tts_kim, rf_inversion} or noising-and-denoising to perturb samples \cite{tts_ma}, incur large computational overheads:
SDE samplers typically require many more steps \cite{songscore, dpm, flow_gen, flow_fast, tts_kim}, and every additional denoising step reduces TTS efficiency.
As a result, local trajectory methods that exhibit excellent performance for diffusion models \cite{svdd, code, tts_ma, tts_greedy, fk, classic} underperform for flow matching
models, often worse than simple best-of-$N$ for the same computational budget.
Following this observation, we opt to eschew expensive local search and instead allocate the computational budget to explore the largest possible pool of diverse global candidates, without fully denoising all candidates.

\begin{figure}[!t]\centering
  \begin{minipage}[c]{0.42\linewidth}\centering
        \begin{tikzpicture}[scale=0.7]
    \begin{axis}[
        ybar,
        bar width=8pt,
        width=7.5cm,
        height=5.2cm,
        enlarge x limits=0.08,
        ylabel={LPIPS},
        ymin=0.35,
        ymax=0.75,
        ytick={0.40, 0.50, 0.60, 0.70, 0.80},
        xtick=data,
        xticklabels={
            Flux2-klein,
            Flux1.dev,
            Pixelart-$\Sigma$,
            HunyuanDiT,
            Playground~2.5,
            SDXL,
            SD3.5-L,
            SD3-M,
            SD2
        },
        xticklabel style={
            rotate=35,
            anchor=north east,
            font = \small
        },
        tick align=outside,
        axis x line*=bottom,
        axis y line*=left,
        ymajorgrids=true,
        grid style={dotted, gray!50},
    ]
    \addplot[fill=blue!40!white, draw=blue!60!black] coordinates {
        (1, 0.4960)
        (2, 0.5366)
        (3, 0.5553)
        (4, 0.5725)
        (5, 0.6058)
        (6, 0.6166)
        (7, 0.6322)
        (8, 0.6325)
        (9, 0.7187)
    };
    \end{axis}
\end{tikzpicture}
        \caption{Motivation of Repel:  Diversity on HPS-test set measured via averaged LPIPS distances between multiple same-prompt image pairs. 
      Better-scaled diffusion \& flow models exhibit lower diversity. This increases the importance of the Repel mechanism for TTS.
        }
        \label{fig:diversity_bar}
    \end{minipage}\hfill
    \begin{minipage}[c]{0.56\linewidth}
        \begin{subfigure}[c]{\linewidth}\centering
            \includegraphics[width=\linewidth]{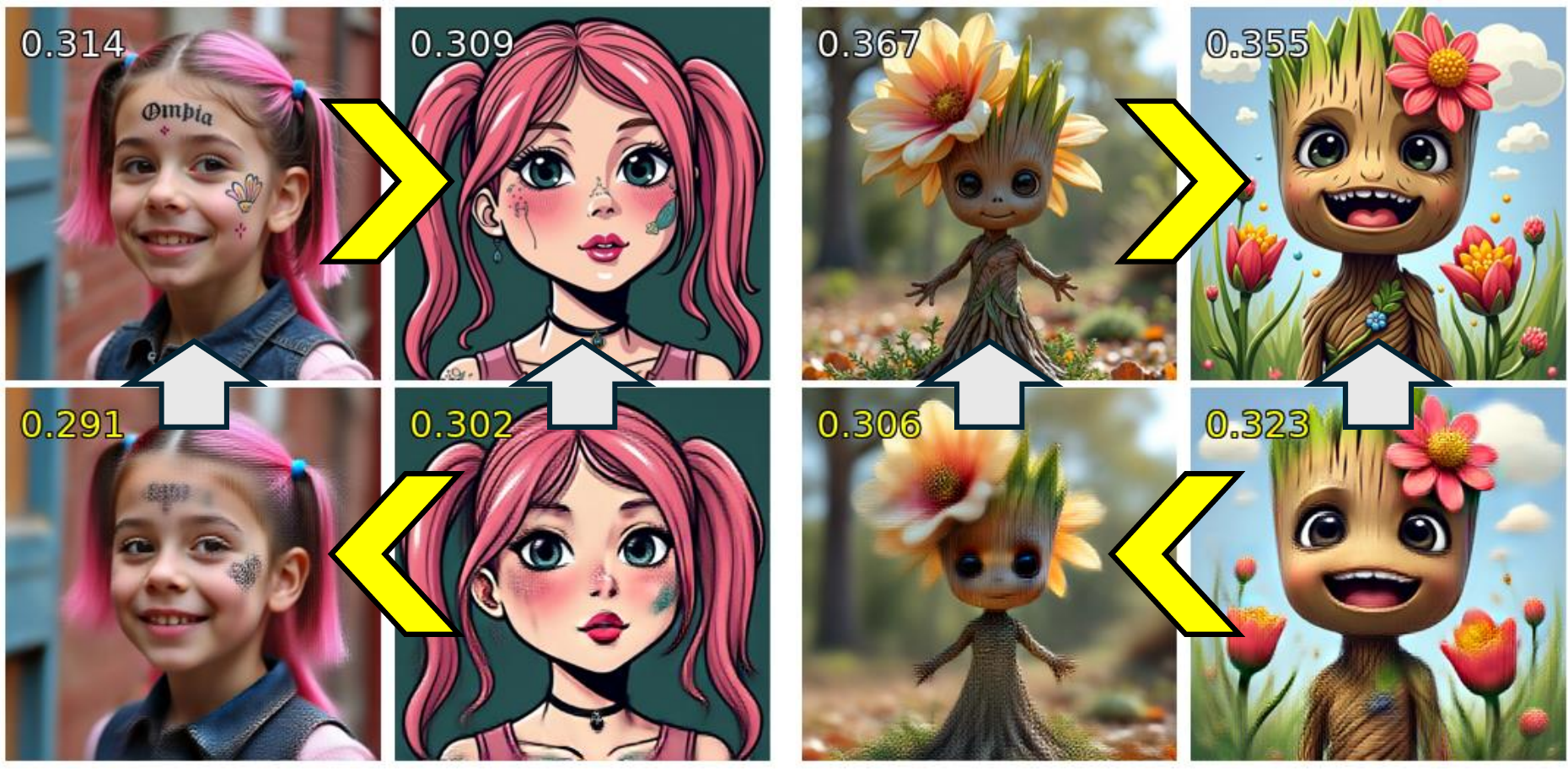}
        \end{subfigure}
        \caption{ 
       Motivation of NARF:
      (Top) Fully denoised images. (Bottom) Partially denoised (60\%) estimations. 
      Reward models trained on clean data misrank partially denoised samples, where high-frequency details are not fully recovered.  
      As a consequence, this would favor "cartoonish" images, causing high mis-pruning rates. 
      NARF helps correct this bias.}
    \end{minipage}
    \label{fig:bias}
\end{figure}

Specifically, we propose a straightforward global pruning strategy that generates numerous candidates in parallel, verifies them at intermediate steps, and discards low-scoring samples.
However, this introduces two critical challenges: limited candidate diversity and poor reward model performance at intermediate denoising steps.
First, our performance improvements stem from exploring a large pool of candidates under a fixed computation budget.
Modern flow matching models are typically trained on very large datasets, but tend to converge to a sharp, high-density region of the data manifold \cite{model_collapse, diversity_gass, negtome, collaps_recovery}.
When the search bandwidth is limited, increasing the number of searching candidates yields diminishing returns.
Second, flow models denoise progressively from coarse, low-frequency structures to fine, high-frequency details.
Consequently, relying on off-the-shelf image reward models that were trained exclusively on clean images for verifying noisy images leads to poor TTS performance and a high-frequency bias, \eg, preferring cartoonish images, as shown in \cref{fig:bias}.

To address these limitations, we propose two efficient methods.
First, to improve candidate diversity, we introduce Repel, a token-repelling mechanism inspired by \cite{negtome} that pushes apart similar tokens across different trajectories during generation, while requiring negligible additional computation.
Second, to ensure accurate early pruning, we bridge the reward model domain gap between clean and noisy images by introducing NARF, noise-aware reward fine-tuning via self-distillation to produce reliable estimates at different noise levels.
Specifically, we generate training data automatically by using the final clean-image reward to supervise intermediate Euler estimates.
By adopting a curriculum that progresses from near-clean to increasingly noisy timesteps, this strategy achieves stable convergence.
Overall, we summarise our TTS pipeline as \textbf{DOG-Trim}: \textbf{D}iversity enhanced \textbf{O}rder aligned \textbf{G}lobal flow Trimming.
Our contributions are
\begin{enumerate}
    \item a global TTS search strategy that we show is more suitable for flow matching models than local or hybrid approaches, while elucidating its key challenges;
    \item a token-level repulsion mechanism that enhances generative diversity across parallel trajectories, expanding the effective search space; and
    \item an efficient noise-aware reward self-distillation strategy for extending reward models to the domain of noisy images.
\end{enumerate}

\section{Method}
\label{sec:method}

\subsection{Preliminaries}

\subsubsection{Flow-Matching Models.}
Flow matching learns the probability flow between prior and target data distributions \cite{edm, dpm, flow_gen, flow_fast}.
In the forward process, the conditional distribution of the latent at timestep $t \in [0,1]$ is formulated as
\begin{align}
     p( x_{t} |x_{0}) =\mathcal{N}\left( x_{0} ,\sigma(t)^{2} I \right),
\end{align}
for a clean latent $x_0$ and noise schedule $\sigma(t)$.
The reverse process is governed by a pretrained neural network with parameters $\theta$ trained to estimate the flow velocity $v_\theta(x_t)$.
The two primary forms defining this process are
\begin{align}
\text{ODE:} \quad dx_t &= v_\theta(x_t) dt \\
\text{SDE:} \quad dx_t &= \left( v_\theta(x_t) - \frac{g_t^2}{2} \nabla \log p_\theta(x_t) \right) dt + g_t dW_t,
\end{align}
where the ODE describes a deterministic trajectory, and the SDE defines a stochastic sampling path via the Wiener process $W_t$, a compensating predefined term $g_t$, and the score $\nabla \log p_\theta(x_t)$, which can be computed from the velocity.
Although the SDE and ODE describe the same marginal distribution of $x_t$, ODEs have become the standard choice in modern large-scale flow-matching models for text-to-image synthesis \cite{flux, sd3, sana, qwenimage}, as they require fewer inference steps and converge more reliably to the data manifold \cite{dpm, dpm++}.

\subsubsection{TTS for Diffusion and Flow Models.}
The inference cost of text-to-image diffusion or flow models can be naturally scaled by increasing the number of denoising steps, but this yields rapidly diminishing returns \cite{tts_ma}.
A more effective strategy is to search over a pool of sampled candidates to identify the one that leads to the highest quality generation \cite{tts_ma, fk, classic, tts_kim, sana1_5, das}.
This process relies on diverse samples and two key components: a verifier and a search algorithm.
Verifiers are typically pretrained lightweight reward models \cite{imagereward, hps, pickapic, clip} that evaluate the quality of generated image candidates.
Conditioned on an input text prompt $p \in \mathcal{P}$, the verifier is formulated as
\begin{align}
      \psi:
      \mathbb{R}^{H \times W \times C} \times \mathcal{P} \to \mathbb{R},
\end{align}
and may not be differentiable with respect to the inference trajectory \cite{das}.
The search algorithm $f$ aims to identify the most promising candidates that maximize the verifier's reward scores.
It takes the verifier $\psi \in \Psi$, the flow-matching model $\phi \in \Phi$, and $N$ candidate image samples with text conditions as input and returns an image,
\begin{align}
    f: \Psi \times \Phi \times \mathbb{R}^{N \times H \times W \times C} \times \mathcal{P}  \to \mathbb{R}^{H \times W \times C}. \label{eq:tts_alg}
\end{align}
Increasing the number of candidates $N$ increases the probability of finding a high-reward sample, but also increases the inference-time compute cost.
Therefore, the critical challenge in designing an effective algorithm $f$ is to maximize the search space while minimizing the computational overhead.
While some methods scale test-time performance by iteratively refining the conditioning signals \cite{reflect_perfect, reflectdit, tts_kim_p}, our study specifically focuses on optimizing the search process under a fixed text condition.

\subsection{Global Search for Test-Time Scaling of Flow Models}

Rather than fully denoising initial noise samples, as done in the best-of-$N$ approach, recent methods improve efficiency through local trajectory-level steering \cite{tts_greedy, das, fk, classic, tts_kim, tts_tree}.
Despite their varying designs, the core principle remains consistent: under a fixed computation budget, prefer to explore locally in more promising regions.
While integrating local exploration can enhance search efficiency, these trajectory-level designs are not inherently compatible with flow-matching models.
Although both diffusion and flow paradigms denoise from a Gaussian prior, flow models generally rely on deterministic ODE solvers.

To enable local trajectory branching, one alternative is to replace the ODE sampler with an SDE.
However, SDEs typically require far more denoising steps than ODEs, as indicated in prior work \cite{dpm++, flow_gen, flow_fast}.
To empirically validate this, we conducted a preliminary experiment on the HPS test set \cite{hps}, where we progressively increase the number of SDE sampling steps \cite{rf_inversion}.
We observe that even when allocating three times the number of timesteps used by our ODE baseline, the SDE's reward values still exhibited a substantial performance gap compared to the ODE, as shown in \cref{fig:sde}.
Another approach is to resample candidates along the ODE trajectory using a noise-then-denoise mechanism \cite{tts_ma}.
However, this requires more denoising steps, which incurs substantial computational overhead.
In summary, there is no free lunch when it comes to local trajectory search for flow models.
In practice, the computation cost of these alternatives completely outweighs the benefits of local exploration, as we demonstrate in \cref{tab:main}.

Based on these findings, we propose a simple but effective global triming strategy that is well-suited for ODE-based flow-matching models.
Starting with $N$ initial candidates, we denoise them in parallel to a predefined intermediate timestep $t$.
For verification, we compute the clean Euler estimates $x_{0|t} = x_t + v_\theta t$, decode them to image space, compute their image rewards, and prune the lowest-scoring trajectories.
The surviving candidates are denoised further, verified, pruned, and so on, until the remaining candidates are fully denoised.
This set is decoded into image space, their image rewards are computed, and the image with the greatest reward is selected.
For the pruning stage, we use a fixed retain ratio hyperparameter $\gamma \in [0, 1]$, and keep the top candidates of this fraction.
Prior works also leverage adaptive re-sampling based on the score distribution of all candidates.
For comparison, we adapt this approach to our pruning setting and discuss the differences in \cref{exp:ablation}.
Both strategies have comparable performance under the same computation budget, but the fixed threshold approach makes it significantly easier to pre-allocate resources given a specific computation budget.

\subsection{Enhancing Sample Diversity with Repel}

\begin{figure}[!t]\centering
     \begin{subfigure}[]{\linewidth}\centering
        \includegraphics[width=\linewidth]{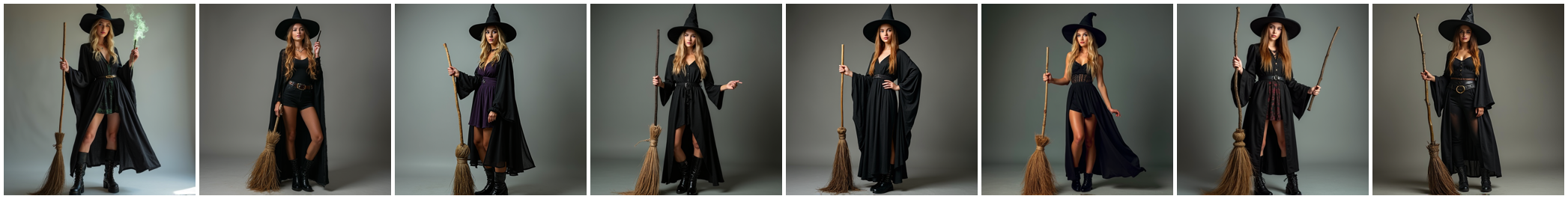}
        \includegraphics[width=\linewidth]{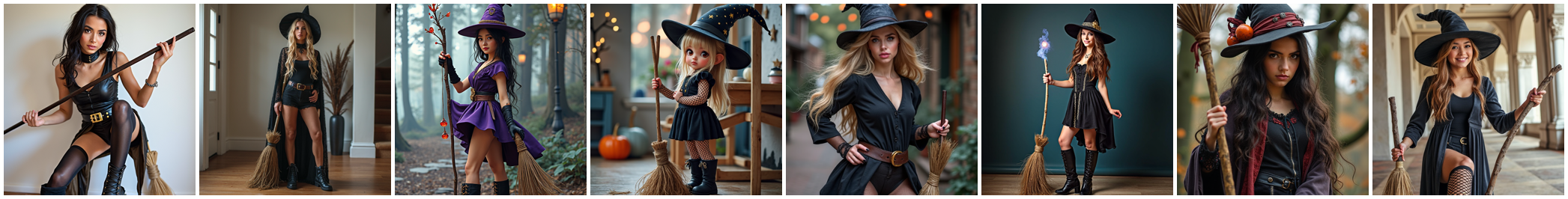}
        \caption{Prompt: \textit{A young woman witch cosplaying with a magic wand and broom, wearing boots, and posing in a full body shot with a detailed face.}}
    \end{subfigure}
     \begin{subfigure}[]{\linewidth}\centering
        \includegraphics[width=\linewidth]{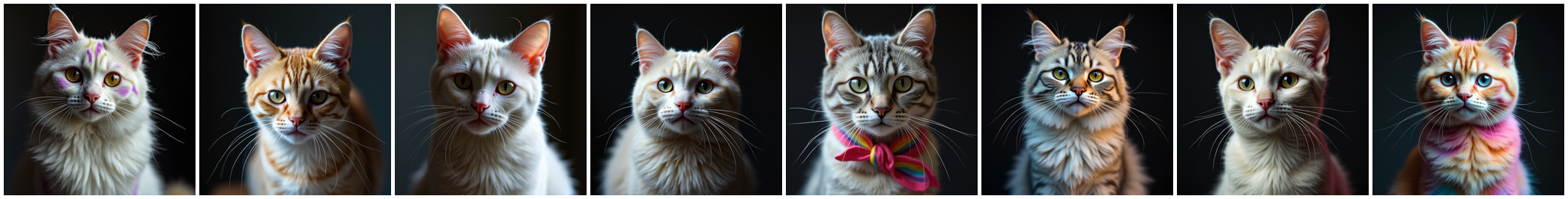}
        \includegraphics[width=\linewidth]{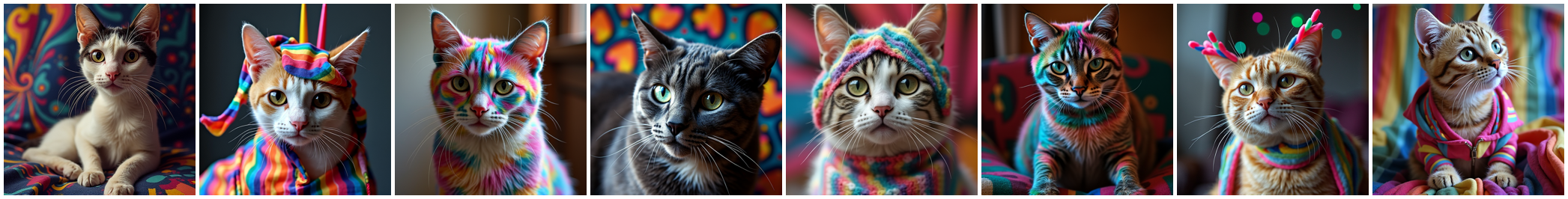}
         \caption{Prompt: \textit{A portrait of Nyan Cat, styled after Annie Leibovitz's dramatic photography.}}
    \end{subfigure}
     \begin{subfigure}[]{\linewidth}\centering
        \includegraphics[width=\linewidth]{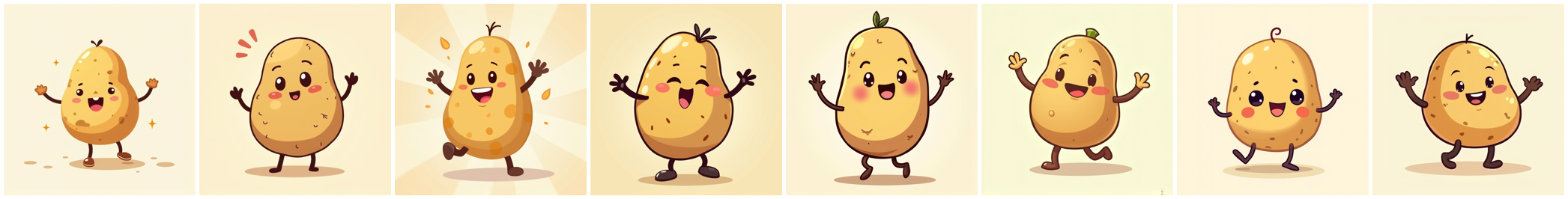}
        \includegraphics[width=\linewidth]{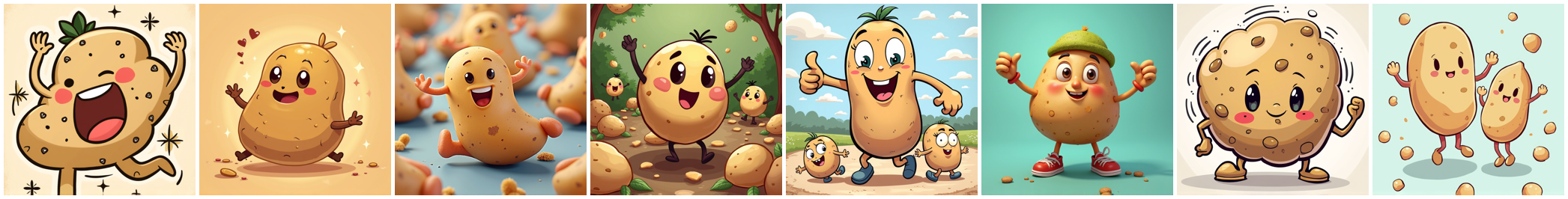}
        \caption{Prompt: \textit{The image is of dancing potatoes in a cute cartoony style.}}
    \end{subfigure}
    \caption{
        Generated images without Repel (top row) and with Repel (bottom row).
    }
    \label{fig:token_repel}
\end{figure}

As discussed in \cref{sec:intro}, flow-matching models tend to produce similar outputs for a given prompt, heavily restricting the effective search space.
Inspired by Singh \etal~\cite{negtome}, we introduce Repel, a cross-trajectory token-repelling mechanism that enhances the diversity of samples generated in parallel.
The core idea is to actively repel each output token away from its most similar counterpart of other images within the batch, thereby forcing the generation paths to diverge.
This operation is applied to the image tokens of MMDiT blocks~\cite{dit, sd3}.

Concretely, let $\mathbf{O} \in \mathbb{R}^{B \times N \times D}$ denote the token features after a transformer block across $B$ parallel trajectories, where $N$ is the number of image tokens and $D$ is the feature dimension.
We flatten this tensor to $\bar{\mathbf{O}} \in \mathbb{R}^{BN \times D}$ and compute the pairwise cosine similarity matrix, explicitly masking out intra-trajectory (same-image) pairs.
This process is formulated as
\begin{align}
    \mathbf{S} &= \bar{\mathbf{O}}_{\text{norm}} \, \bar{\mathbf{O}}_{\text{norm}}^\top, \; \mathbf{S}_{ij}=-\infty \; \text{if} \; i \equiv j \bmod N,  \label{eq:sim} \\
   \mathbf{H} &= \mathbf{1}[\max_j \mathbf{S}_{ij} > \delta], \label{eq:mask}  \\
   \bar{\mathbf{O}}' \!&= ((1+\alpha)\bar{\mathbf{O}} - \alpha\,\bar{\mathbf{O}}[\arg\max_j \mathbf{S}_{ij}]) \odot \mathbf{H} + \bar{\mathbf{O}} \odot (1 - \mathbf{H}).
    \label{eq:push}
\end{align}
First, in \cref{eq:sim}, we compute the cosine similarity between each token and all tokens from other images in the batch, where $\bar{\mathbf{O}}_{\text{norm}}$ denotes $\bar{\mathbf{O}}$ normalized along the feature dimension.
Second, in \cref{eq:mask}, we identify the most similar cross-trajectory correspondent for each token and compute a binary mask $\mathbf{H}$, thresholded by a hyperparameter $\delta$ to isolate only highly correlated pairs.
Then, in \cref{eq:push}, we linearly extrapolate the selected tokens away from their closest matches, effectively increasing the inter-trajectory feature distance, with $\alpha > 0$ controlling the repulsion strength.
Finally, the modified tensor $\bar{\mathbf{O}}'$ is reshaped back to $\mathbb{R}^{B \times N \times D}$ and passed into the MLP as usual.
Crucially, this mechanism is only applied during the initial few denoising steps, when the global semantic structure is being established and diversity amplification yields the highest impact.
In the later stages of generation, applying feature repulsion can introduce visual artifacts and degrade image quality.

\subsection{Noise-Aware Image Rewards for Look-Ahead Search}
\label{subsec:reward}

As discussed in \cref{sec:intro}, applying an off-the-shelf reward model to the Euler estimates of intermediate (noisy) images has a severe domain gap issue.
Standard reward models are trained exclusively on clean images; when evaluated on early-timestep images, they yield rankings inconsistent with the clean-image rankings.
In particular, they are heavily biased towards high-frequency detail, which for intermediate images tends to be simple and cartoonish images with strong edges, as shown in \cref{fig:bias}.
In contrast, lower-frequency intermediate images are down-ranked, even though they may be much more detailed and diverse once fully denoised.
To address this, we introduce NARF, noise-aware reward finetuning via self-distillation, enabling the reward model to ``look ahead'' and accurately predict the clean image reward from early, noisier generation states.

\paragraph{Pipeline design.}
To align the intermediate reward with the final reward, we propose a self-distillation pipeline.
Training data is generated autonomously by sampling trajectories from the pretrained flow model across various random noise initializations.
At each target intermediate timestep, we extract the image representation via decoded Euler estimates on the latent state.
The pseudo-ground-truth training target for these intermediate states is simply the score assigned to the final, fully-denoised image of that exact trajectory by a pretrained reward model.
The noise-aware model is then trained via a simple MSE loss to regress this final target score.
This design inherently captures the true marginal distribution of intermediate states encountered during inference, eliminating the need for large-scale training datasets.
Furthermore, this pipeline is highly scalable, requiring only a prompt list and no human data curation.

\paragraph{Training strategies.}
To further boost the efficiency of this distillation process, we incorporate two key training strategies: diversity augmentation and curriculum learning.
First, standard reward models rely on training batches with distinct quality variations to learn meaningful rankings~\cite{imagereward, hps, pickapic}.
However, parallel trajectories conditioned on the same prompt often collapse into highly similar outputs, limiting the variance necessary for effective regression.
To mitigate this, we inject the aforementioned cross-trajectory repulsion mechanism Repel during data generation, pushing the samples apart to dynamically construct diverse training batches.
Second, regressing the final reward from highly noisy, early-timestep images is fundamentally difficult due to severe information loss.
Therefore, we adopt an easy-to-hard curriculum training strategy.
We initialize the finetuning on near-clean images at late timesteps and progressively shift the training domain backward toward noisier intermediate states, saving the model weights at each target time-step.
By keeping the domain shift small, this approach ensures high training efficiency and stable convergence at each timestep.

\section{Experiments}
\label{sec:experiment}

\subsection{Experiment Setup}
\label{exp:setup}

We adopt Flux.1-dev~\cite{flux} as our base flow-matching model, configured with 35 inference steps and a guidance scale of 3.5.
For Repel, we set the repulsion strength $\alpha = 3.0$, the similarity threshold $\delta=0.7$, and the token dropout rate to 0.2, applying it exclusively during the early sampling stages (from timestep 1000 down to 900) across all MMDiT blocks.
Notably, we apply these exact same settings for both training data augmentation and test-time scaling.
We test our method with four different reward models as our verifier: ImageReward~\cite{imagereward}, HPSv2.1~\cite{hps}, PickScore~\cite{pickapic}, and a reward ensemble that sums the rank of the other three rewards.
Each of these have been finetuned for different noise levels, as outlined in the previous section.
To construct the training dataset, we select the top 5,000 prompts from the HPD training split~\cite{hps} and generate 4 parallel candidate trajectories per prompt.
During inference, the pruning stages are uniformly distributed at 20\% intervals across the denoising process, occurring every 7 steps.
For candidate evaluation, we use the original pretrained reward models at timesteps 28 and 35, and deploy our finetuned models for the intermediate timesteps 21, 14, and 7.

\subsection{Results}
\label{exp:main}

We compare with existing methods under the same computation budget, equivalent to fully-denoising 6 images, and evaluate using all three image rewards.
Since most existing test-time scaling methods are designed for stochastic diffusion models,
we adapt an SDE sampler \cite{rf_inversion} on Flux to facilitate comparison.
The following baselines are considered.

\paragraph{Local Baselines.}
First, we implement Greedy and $\epsilon$-Greedy Search \cite{tts_ma, tts_greedy} to explore the initial noise space.
We iteratively sample $m=2$ neighbors around the best-performing noise for $k=3$ iterations, with $\epsilon=0.4$ in $\epsilon$-Greedy.
Second, we evaluate a trajectory-level local baseline Traj-Greedy \cite{svdd, code} that begins with a single candidate and explores its $m$ neighbors at each 20\% denoising interval with SDE.
Finally, we implement an ODE-based trajectory local search called Search-over-Path (SoP) \cite{tts_ma}, where we evaluate intermediate states every 10 denoising steps, select the best candidate, inject 5 steps of noise, and resample for 10 steps.
To strictly adhere to the computation budget, we keep exactly 4 active candidates per iteration.

\paragraph{Hybrid Baselines.}
Sequential Monte Carlo (SMC) based methods integrate local and global exploration with a beam search structure: denoising all candidates to the target intermediate stages and then resampling them based on their reward distribution.
First, we compare our approach against a representative work, FK-Steering \cite{fk}.
Second, we compare with the improved SMC approach in Zhang \etal~\cite{classic}, which we denote iSMC.
We strictly follow their default settings, modifying only the tempering mechanism used to compute normalized reward weights at each verification step.
Since the tempering hyperparameter $\lambda$ depends heavily on the scale of the reward, we performed a coarse hyperparameter sweep on a validation set, ultimately setting $\lambda = 50$, $0.02$, $0.008$, and $0.001$ for HPSv2.1, ImageReward, PickScore, and the Ensemble, respectively.
The number of initial candidates are set at 6.

\paragraph{Global Baselines.}
Other than the classic global method best-of-$N$ (BoN) \cite{tts_ma, sana1_5} with $N=6$, we adapt several trajectory search algorithms into adaptive pruning strategies under deterministic ODE sampling.
For Trajectory Greedy, we convert it to a fixed-stage early selection algorithm: we sample $N$ initial candidates, denoise them to $40\%$ and select the best.
We increase the number of candidates to 13 to match the computation budget.
For FK-Steering and iSMC, we apply the `unique' operation after local resampling.
This allows the methods to adaptively set the number of active samples, depending on the sharpness of the distribution.
We empirically ascertained that the budget was approximately satisfied by setting $N$ to 8, 10, and 12 for the HPSv2.1, ImageReward, PickScore, and Ensemble reward models.

\begin{figure}[!t]
    \centering
    \begin{subfigure}{\linewidth}
        \centering
        \makebox[0.02\linewidth][c]{\raisebox{1em}{\rotatebox{90}{\small\textbf{Ours} \quad\quad\quad iSMC \quad\quad\quad T-Greedy \hspace{2pt} \quad\quad \  SoP \quad\quad\quad \ Greedy \quad \quad \quad BoN \quad\quad \ Random}}}
        \includegraphics[width=0.13\linewidth]{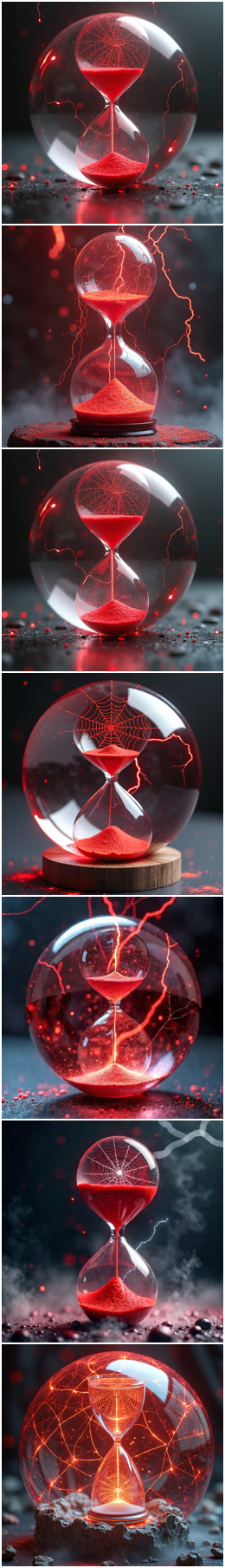}\hfill
        \includegraphics[width=0.13\linewidth]{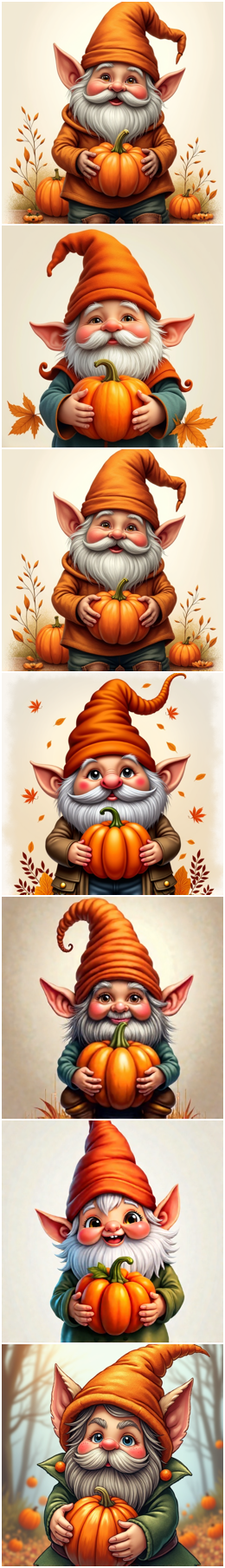}\hfill
        \includegraphics[width=0.13\linewidth]{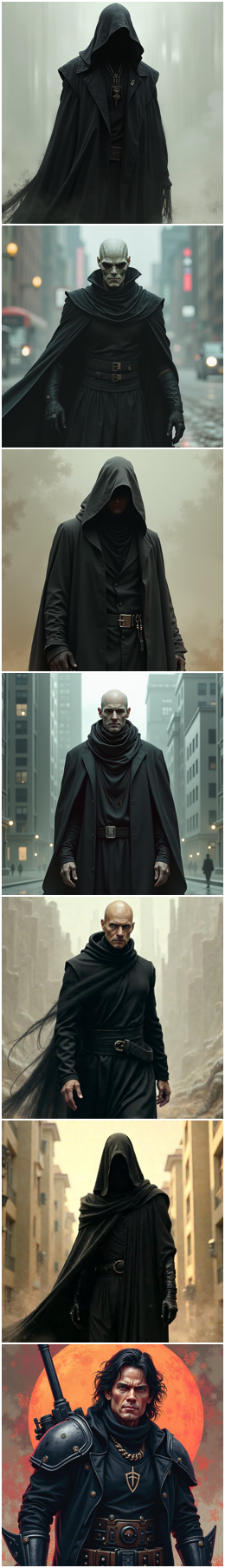}\hfill
        \includegraphics[width=0.13\linewidth]{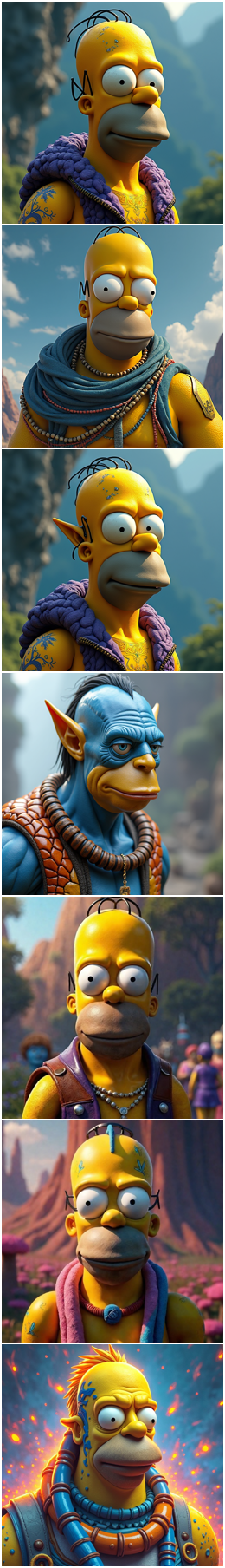}\hfill
        \includegraphics[width=0.13\linewidth]{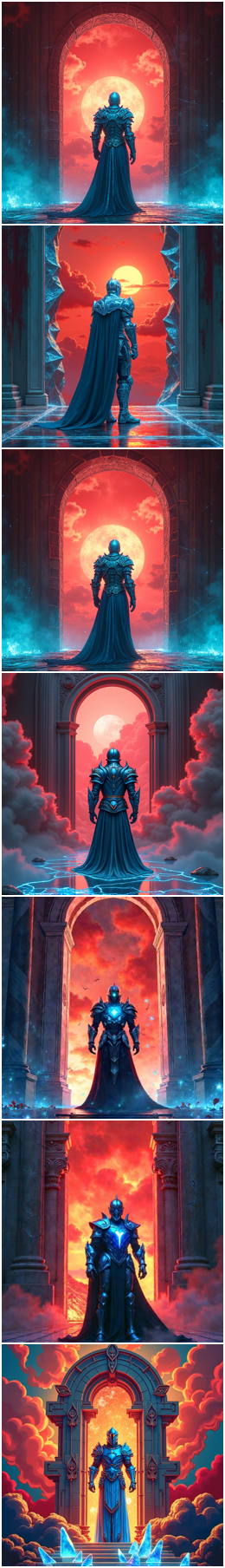}\hfill
        \includegraphics[width=0.13\linewidth]{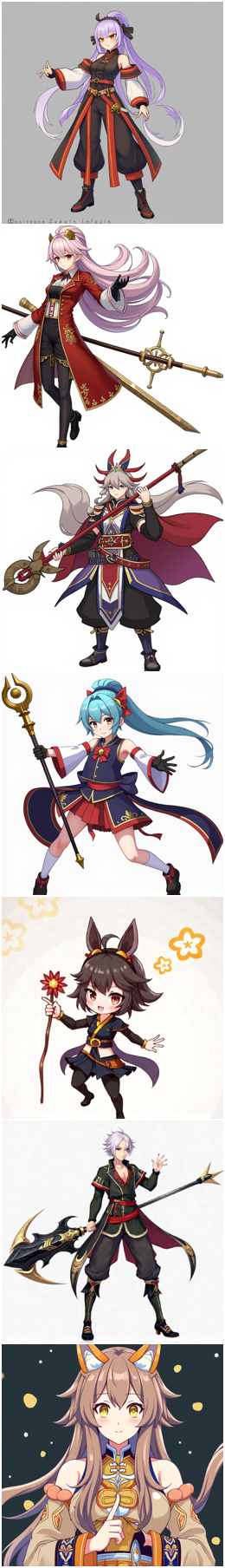}\hfill
        \includegraphics[width=0.13\linewidth]{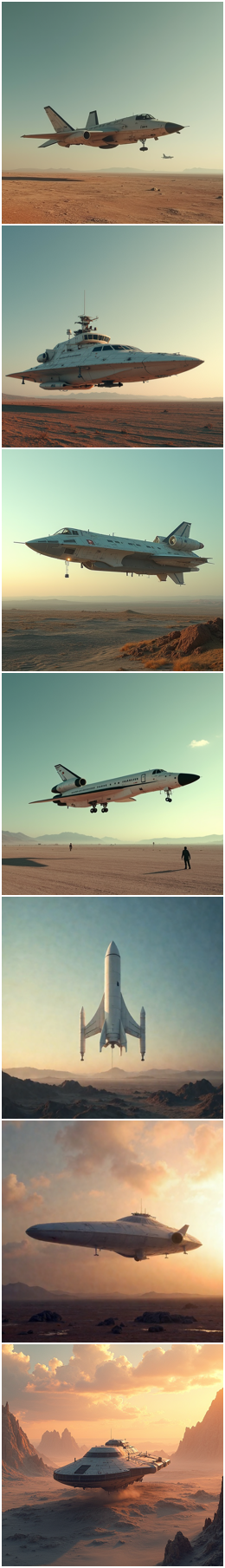}
    \end{subfigure}
    \caption{
        Qualitative results comparing TTS methods.
        Prompts from left to right:
        \textit{
        \hlc[pink!30]{1.} A red cobweb is seen inside a marble with an hourglass, lightning and intricate details, creating a sense of awe with swirling mist;
        \hlc[yellow!40]{2.} A hand-drawn cute gnome holding a pumpkin in an autumn disguise, portrayed in a detailed close-up of the face with warm lighting and high detail;
        \hlc[green!20]{3.} Sandman wearing black clothing, in a sci-fi themed digital painting by Greg Rutkowski;
        \hlc[cyan!20]{4.} The image is a portrait of Homer Simpson as a Na'vi from Avatar, created with vibrant colors and highly detailed in a cinematic style reminiscent of romanticism by Eugene de Blaas and Ross Tran, available on Artstation with credits to Greg Rutkowski;
        \hlc[pink!30]{5.} A warrior in glowing azure plate armor stands in a doorway to hell sliced by iridescent glass cracks, with crimson clouds and an art deco palace backdrop;
        \hlc[yellow!40]{6.} Keqing from Genshin Impact;
        \hlc[green!20]{7.} A spaceship in an empty landscape.
        }
    }
    \label{fig:qualitative}
\end{figure}

\paragraph{Findings.}
Based on the results from \cref{tab:main}, we can observe that
(1)~out of the methods that search the noise prior directly, best-of-$N$ performs better than Greedy and $\epsilon$-Greedy;
(2)~SDE-based hybrid search methods consistently underperform their ODE and pruning-based counterparts for the same computation budget; and
(3)~our method significantly outperforms the other approaches, on all metrics.
From findings 1 and 2, we can see that the additional compute cost associated with local search for flow-matching models outweighs the benefits: FLOP-for-FLOP, it is better to just add more initial global candidates.
Qualitative results are presented in \cref{fig:qualitative}, where it is clear that our approach generates better quality images with greater text prompt fidelity.

\begin{table}[!t]
    \centering
    \small
    \setlength{\tabcolsep}{3pt}
    \caption{
    Quantitative comparison of our proposed approach and existing test-time scaling methods.
    We report the average final reward values under both single-reward and combined-reward (ensemble) settings.
    Every method has a compute budget approximately equivalent to fully-denoising six images.
    }
    \label{tab:main}
    \renewcommand{\arraystretch}{1.1}
    \begin{tabularx}{\linewidth}{@{} lc *{3}{>{\centering\arraybackslash\hsize=0.75\hsize}X} >{\centering\arraybackslash\hsize=1.75\hsize}X @{}}
        \toprule
        \multicolumn{2}{r}{Verifier Reward:} & HPS & Pick & ImR & Ensemble \\
        \multicolumn{2}{r}{Evaluation Reward:} & HPS & Pick & ImR & HPS / Pick / ImR \\
        \midrule
        Random  & ODE & 0.3029 & 22.61 & 0.994 & 0.3029 / 22.61 / 0.994 \\
        BoN & ODE & 0.3211 & 23.26 & 1.385 & 0.3173 / 23.13 / 1.301 \\
        Greedy & ODE & 0.3178 & 23.17 & 1.308 & 0.3107 / 22.90 / 1.158 \\
        $\epsilon$-Greedy & ODE & 0.3194 & 23.26 & 1.360 & 0.3115 / 22.92 / 1.182 \\
        SoP & ODE & 0.3197 & 22.94 & 1.322 & 0.3167 / 22.77 / 1.209 \\
        Traj-Greedy & SDE & 0.3118 & 22.84 & 1.341 &  0.3060 / 22.59 / 1.175 \\
        Traj-Greedy & ODE  & 0.3184 & 23.14 & 1.376 & 0.3165 / 23.02 / 1.291 \\
        FK-Steer & SDE & 0.3133  & 22.71  & 1.336 & 0.3040 / 22.38 / 1.136 \\
        FK-Steer & ODE & 0.3161 & 23.19 & 1.331  &  0.3153/ 23.09/ 1.268 \\
        iSMC & SDE & 0.3154  & 22.86  & 1.333 & 0.3080/ 22.66/ 1.229 \\
        iSMC & ODE & 0.3225 & 23.11 & 1.365 & 0.3144/ 22.95/ 1.247 \\
        Ours & ODE & \textbf{0.3314}  & \textbf{23.35} & \textbf{1.682} & \textbf{0.3268}/ \textbf{23.14}/ \textbf{1.579} \\
        \bottomrule
    \end{tabularx}
\end{table}

\subsection{Ablation Study and Analyses}
\label{exp:ablation}

\paragraph{Ablation Study.}
As detailed in \cref{tab:ablation_comp}, we evaluate the individual contributions of global pruning, token repulsion (Repel), and noise-aware reward finetuning (NARF).
Global pruning alone establishes a strong foundation that already outperforms prior state-of-the-art methods.
Building on this, Repel and NARF both independently improve the average attained reward.

\begin{table}[!t]
    \centering
    \small
    \renewcommand{\arraystretch}{1.1}
    \setlength{\tabcolsep}{4pt}
    \caption{Ablation study of primary components, where first row represents no-TTS baseline and second row represents the outcome of the best baseline method.}
    \label{tab:ablation_comp}
    \begin{tabularx}{\linewidth}{@{} ccc CCC @{}}
        \toprule
        Global Prune & Repel & NARF & HPSv2.1 $\uparrow$ & PickScore $\uparrow$ & ImageReward $\uparrow$  \\
        \midrule
        \textit{Random} & &     & 0.3114 & 22.89 & 1.163 \\
        \textit{Best}   & & & 0.3331 & 23.59 & 1.534\\
        $\checkmark$ &              &              & 0.3338 & 23.63 & 1.582 \\
        $\checkmark$ & $\checkmark$ &              & \underline{0.3409} & 23.56 & 1.603 \\
        $\checkmark$ &              & $\checkmark$ & 0.3349 & \textbf{23.72}  & \underline{1.820} \\
        $\checkmark$ & $\checkmark$ & $\checkmark$ & \textbf{0.3443} & \underline{23.67} & \textbf{1.843} \\
        \bottomrule
    \end{tabularx}
\end{table}

\paragraph{On Repel.}
Token repulsion effectively increases the diversity of the sampled images.
To validate this, we generated eight images for each of the 100 prompts in our validation set, analyzing the mean and standard deviation of their rewards across different values of the repulsion hyperparameter $\alpha$ (\cref{eq:push}).
Setting $\alpha=0$ reduces the process to the original Flux model, while increasing $\alpha$ yields higher diversity.
As shown in \cref{plot:token_repel}, increasing $\alpha$ yields a wider distribution of the reward scores, but too large an $\alpha$ degrades the average image quality by pushing tokens out of distribution.
Based on this evidence, we selected $\alpha$ as $3.0$.
Qualitative results of Repel under this setting is shown in \cref{fig:token_repel}.

\begin{figure}[!t]
    \centering
    \begin{subfigure}{0.32\textwidth}
        \centering
        \begin{tikzpicture}[scale=0.7]

\definecolor{myyellow}{RGB}{255, 180, 60}
\definecolor{myred}{RGB}{245, 70, 100}
\definecolor{myblue}{RGB}{70, 150, 220}
\definecolor{mygreen}{RGB}{60, 200, 110}

\colorlet{mybluelight}{myblue!30!white} 

\begin{axis}[
    width=6cm, 
    height=4cm,
    axis x line*=bottom,
    axis y line*=left,
    legend cell align={left},
    legend style={
        draw=none,
        at={(1.0, 0.05)},        
        anchor=south east,
        legend columns=1,
        fill=none,
        font=\small
    },
    xtick style={black},
    ytick style={black},
    xtick distance=1,     
    ytick distance=0.02,  
    ymin=0.25,            
    ymax=0.33,
    xmin= -0.5,            
    xmax= 5.5,             
    tick align=outside,
    tick pos=left,
    tick label style={font=\small},
    xlabel={$\alpha$},
    ylabel={},
    xticklabel style={
        /pgf/number format/fixed,
        /pgf/number format/precision=1
    },
    yticklabel style={
        /pgf/number format/.cd,
        fixed,
        fixed zerofill,
        precision=2,
    },
    axis lines = left,
]

\addplot [name path=upper, draw=none] table [x index=0, y index=2] {
    0.0 0.3000 0.3128 0.2872
    1.0 0.2975 0.3130 0.2820
    2.0 0.2976 0.3157 0.2795
    3.0 0.2985 0.3179 0.2791
    4.0 0.2952 0.3172 0.2732
    5.0 0.2867 0.3127 0.2607
};

\addplot [name path=lower, draw=none] table [x index=0, y index=3] {
    0.0 0.3000 0.3128 0.2872
    1.0 0.2975 0.3130 0.2820
    2.0 0.2976 0.3157 0.2795
    3.0 0.2985 0.3179 0.2791
    4.0 0.2952 0.3172 0.2732
    5.0 0.2867 0.3127 0.2607
};

\addplot [fill=mybluelight, opacity=0.4] fill between[of=upper and lower];

\addplot [
    color=myblue, 
    line width=2pt, 
    mark=*, 
    mark options={fill=myblue, scale=0.8, solid}
] table [x index=0, y index=1] {
    0.0 0.3000 0.3128 0.2872
    1.0 0.2975 0.3130 0.2820
    2.0 0.2976 0.3157 0.2795
    3.0 0.2985 0.3179 0.2791
    4.0 0.2952 0.3172 0.2732
    5.0 0.2867 0.3127 0.2607
};

\addplot [
    only marks,          
    mark=x,              
    mark options={color=myblue, scale=2.0, thick} 
] table {
    0.0 0.317
    1.0 0.3195 
    2.0 0.3220
    3.0 0.3246 
    4.0 0.3243
    5.0 0.3196
};

\end{axis}
\end{tikzpicture}%
        \label{fig:token_push_hps}
        \caption{HPSv2.1}
    \end{subfigure}\hfill
    \begin{subfigure}{0.32\textwidth}
        \centering
        \begin{tikzpicture}[scale=0.7]

\definecolor{myyellow}{RGB}{255, 180, 60}
\definecolor{myred}{RGB}{245, 70, 100}
\definecolor{myblue}{RGB}{70, 150, 220}
\definecolor{mygreen}{RGB}{60, 200, 110}

\colorlet{myyellowlight}{myyellow!30!white} 

\begin{axis}[
    width=6cm, 
    height=4cm,
    axis x line*=bottom,
    axis y line*=left,
    legend cell align={left},
    legend style={
        draw=none,
        at={(1.0, 0.05)},        
        anchor=south east,
        legend columns=1,
        fill=none,
        font=\small
    },
    xtick style={black},
    ytick style={black},
    xtick distance=1,     
    ytick distance=0.5,   
    ymin=21.1,            
    ymax=23.4,
    xmin= -0.5,            
    xmax= 5.5,             
    tick align=outside,
    tick pos=left,
    tick label style={font=\small},
    xlabel={$\alpha$},
    ylabel={},
    xticklabel style={
        /pgf/number format/fixed,
        /pgf/number format/precision=1
    },
    yticklabel style={
        /pgf/number format/.cd,
        fixed,
        fixed zerofill,
        precision=1,
    },
    axis lines = left,
]

\addplot [name path=upper, draw=none] table [x index=0, y index=2] {
    0.0 22.7192 23.1376 22.3008
    1.0 22.5052 23.0329 21.9775
    2.0 22.4737 23.0404 21.9070
    3.0 22.4705 23.0789 21.8621
    4.0 22.3763 23.0432 21.7094
    5.0 22.1104 22.8761 21.3447
};

\addplot [name path=lower, draw=none] table [x index=0, y index=3] {
    0.0 22.7192 23.1376 22.3008
    1.0 22.5052 23.0329 21.9775
    2.0 22.4737 23.0404 21.9070
    3.0 22.4705 23.0789 21.8621
    4.0 22.3763 23.0432 21.7094
    5.0 22.1104 22.8761 21.3447
};

\addplot [fill=myyellowlight, opacity=0.4] fill between[of=upper and lower];

\addplot [
    color=myyellow, 
    line width=2pt, 
    mark=*, 
    mark options={fill=myyellow, scale=0.8, solid}
] table [x index=0, y index=1] {
    0.0 22.7192 23.1376 22.3008
    1.0 22.5052 23.0329 21.9775
    2.0 22.4737 23.0404 21.9070
    3.0 22.4705 23.0789 21.8621
    4.0 22.3763 23.0432 21.7094
    5.0 22.1104 22.8761 21.3447
};

\addplot [
    only marks,          
    mark=x,              
    mark options={color=myyellow, scale=2.0, thick} 
] table {
    0.0 23.3155
    1.0 23.2469 
    2.0 23.26
    3.0 23.293 
    4.0 23.282
    5.0 23.15
};

\end{axis}
\end{tikzpicture}%
        \label{fig:token_push_pick}
        \caption{PickScore}
    \end{subfigure}\hfill
    \begin{subfigure}{0.32\textwidth}
        \centering
        \begin{tikzpicture}[scale=0.7]

\definecolor{myyellow}{RGB}{255, 180, 60}
\definecolor{myred}{RGB}{245, 70, 100}
\definecolor{myblue}{RGB}{70, 150, 220}
\definecolor{mygreen}{RGB}{60, 200, 110}

\colorlet{mygreenlight}{mygreen!30!white} 

\begin{axis}[
    width=6cm, 
    height=4cm,
    axis x line*=bottom,
    axis y line*=left,
    legend cell align={left},
    legend style={
        draw=none,
        at={(1.0, 0.05)},        
        anchor=south east,
        legend columns=1,
        fill=none,
        font=\small
    },
    xtick style={black},
    ytick style={black},
    xtick distance=1,     
    ytick distance=0.4,   
    ymin=0.1,             
    ymax=1.5,
    xmin= -0.5,            
    xmax= 5.5,             
    tick align=outside,
    tick pos=left,
    tick label style={font=\small},
    xlabel={$\alpha$},
    ylabel={},
    xticklabel style={
        /pgf/number format/fixed,
        /pgf/number format/precision=1
    },
    yticklabel style={
        /pgf/number format/.cd,
        fixed,
        fixed zerofill,
        precision=1,
    },
    axis lines = left,
]

\addplot [name path=upper, draw=none] table [x index=0, y index=2] {
    0.0 0.9608 1.2265 0.6951
    1.0 0.8997 1.2676 0.5318
    2.0 0.8589 1.2644 0.4534
    3.0 0.8896 1.2993 0.4799
    4.0 0.8558 1.2730 0.4386
    5.0 0.7284 1.2639 0.1929
};

\addplot [name path=lower, draw=none] table [x index=0, y index=3] {
    0.0 0.9608 1.2265 0.6951
    1.0 0.8997 1.2676 0.5318
    2.0 0.8589 1.2644 0.4534
    3.0 0.8896 1.2993 0.4799
    4.0 0.8558 1.2730 0.4386
    5.0 0.7284 1.2639 0.1929
};

\addplot [fill=mygreenlight, opacity=0.4] fill between[of=upper and lower];

\addplot [
    color=mygreen, 
    line width=2pt, 
    mark=*, 
    mark options={fill=mygreen, scale=0.8, solid}
] table [x index=0, y index=1] {
    0.0 0.9608 1.2265 0.6951
    1.0 0.8997 1.2676 0.5318
    2.0 0.8589 1.2644 0.4534
    3.0 0.8896 1.2993 0.4799
    4.0 0.8558 1.2730 0.4386
    5.0 0.7284 1.2639 0.1929
};

\addplot [
    only marks,          
    mark=x,              
    mark options={color=mygreen, scale=2.0, thick} 
] table {
    0.0 1.3106
    1.0 1.3712 
    2.0 1.3840
    3.0 1.3939 
    4.0 1.3901
    5.0 1.3894
};

\end{axis}
\end{tikzpicture}%
        \label{fig:token_push_imr}
        \caption{ImageReward}
    \end{subfigure}%
    \caption{Mean ($\bullet$), max ($\times$), and variance (shaded region) of rewards as the repulsion hyperparameter $\alpha$ increases.
    While a larger $\alpha$ induces a higher standard deviation, an excessively large $\alpha$ degrades the average image quality.
    }
    \label{plot:token_repel}
\end{figure}

\paragraph{On Noise-Aware Reward Finetuning.}
To evaluate early-stage ranking agreement, we construct a validation set of 100 unseen prompts, generating 20 samples per prompt along with their corresponding intermediate states.
Our goal is to ensure that reward predictions at early inference stages preserve a ranking similar to the final clean-image counterparts so that the best candidate is not inadvertently pruned.
We quantify this consistency using Kendall's $\tau$ coefficient \cite{kendall}, which measures the ordinal agreement (from $1$ for identical to $-1$ for complete reversal) between early-timestep and final-step reward scores.
It reflects the normalized number of pairwise swaps needed to turn one ranking into the other, independent of the value magnitudes.
Additionally, we compute the pruning recall at each timestep, defined as the probability that the ultimate best candidate is retained within the top-$\gamma$ fraction of samples at this timestep, averaged evenly across $\gamma \in [0.1, 0.9]$.
As demonstrated in \cref{exp:narf}, our noise-aware reward self-distillation significantly improves early-stage ranking agreement across all evaluated timesteps ($t=21, 14, 7$).
Naive training on merely 5000 data samples collected from the flow model itself, as described in \cref{subsec:reward}, yields a massive performance leap over the un-finetuned baselines.
While the primary performance gain stems from the introduction of fine-tuning itself, integrating our proposed Repel data augmentation and curriculum learning components provides further improvements.
This enhanced rank consistency ensures our early-step pruning framework is more reliable.

\begin{table}[!t]
    \centering
    \small
    \renewcommand{\arraystretch}{1.1}
    \setlength{\tabcolsep}{4pt}
    \caption{Ablation study of noise-aware reward finetuning.}
    \label{exp:narf}
    \begin{tabularx}{\linewidth}{@{} ccc CCC CCC @{}}
        \toprule
        & & & \multicolumn{3}{c}{Kendall coefficient $\uparrow$} & \multicolumn{3}{c}{Recall $\uparrow$} \\
        \cmidrule(lr){4-6} \cmidrule(l){7-9}
        Finetuning & Repel & Curriculum & t=21 & t=14 & t=7 & t=21 & t=14 & t=7 \\
        \midrule
                     &              &              & 0.693 & 0.521 & 0.258  & 0.944  & 0.863 & 0.712 \\
        $\checkmark$ &              &              & 0.699 & 0.636 & 0.396  & 0.948  & 0.919 & 0.810 \\
        $\checkmark$ & $\checkmark$ &              & \textbf{0.736} & 0.642 & \textbf{0.417} & \textbf{0.962}  & 0.921 & 0.792 \\
        $\checkmark$ &              & $\checkmark$ & 0.699 & 0.638 & 0.368  & 0.948  & 0.919 & \textbf{0.818} \\
        $\checkmark$ & $\checkmark$ & $\checkmark$ & \textbf{0.736} & \textbf{0.645} & 0.410 & \textbf{0.962} & \textbf{0.922} & 0.812 \\
        \bottomrule
    \end{tabularx}
\end{table}

\paragraph{On the Pruning Heuristics.}
First, we analyse the effect of different fixed pruning ratios and change the initial number of candidates to maintain the same compute budget for fair comparison.
As demonstrated in \cref{fig:fix}, performance peaks when the retention ratio $\gamma$ is between 0.4 and 0.6, while both extremes yield sub-optimal results.
Therefore, we adopt $\gamma = 0.5$ across all experiments.

In addition, existing methods often employ adaptive sampling based on intermediate rewards, such as Srinivasan Sampling Process (SSP) \cite{classic, das} and multinomial sampling \cite{fk}.
Beyond a simple fixed retention ratio, we can also integrate these adaptive strategies into our pruning framework.
To strictly adapt their SSP and multinomial sampling for pruning, we compute normalized weights for all candidates using a tempered softmax over their intermediate rewards, then we draw the same number of samples based on these weights and apply the `unique' operation to remove duplicates.
The softmax temperature progressively increases as denoising advances, reflecting growing confidence in the reward predictions.
Furthermore, we modified the SSP algorithm to support adaptive pruning usage.
Given $N$ candidates with normalized weights $W_i$, the candidate $i$ is deterministically retained if $N W_i \ge 1$; otherwise, it is retained with probability $N W_i$.
We evaluated the performance of these four methods---fixed-ratio (ours), adaptive-pruning (ours), unique SSP and unique multinomial---at a fixed compute budget by selecting a uniform set of candidates and performing temperature hyperparameter $\lambda$ search.
The results are shown in \cref{fig:p}, where all heuristics perform comparably and, notably, the simplest fixed-ratio strategy is competitive.

\begin{figure}[!t]
\centering
\begin{subfigure}[]{0.32\linewidth}
    \centering
    \begin{tikzpicture}[scale=0.7]

\definecolor{myyellow}{RGB}{255, 180, 60}
\definecolor{myred}{RGB}{245, 70, 100}
\definecolor{myblue}{RGB}{70, 150, 220}
\definecolor{mygreen}{RGB}{60, 200, 110}
\definecolor{mypurple}{RGB}{140, 90, 210} 

\begin{axis}[
    width=6cm, 
    height=5cm,
    axis x line*=bottom,
    axis y line*=left,
    legend cell align={left},
    legend style={
        draw=none,
        at={(1.0, 0.05)},        
        anchor=south east,
        legend columns=1,
        fill=none,
        font=\small
    },
    xtick style={black},
    ytick style={black},
    xtick distance=0.1,   
    ytick distance=0.001, 
    ymin=0.3400,          
    ymax=0.3455,
    tick align=outside,
    tick pos=left,
    xlabel={Retention ratio $(\gamma)$}, 
    ylabel={Reward Value},
    xticklabel style={
        /pgf/number format/fixed,
        /pgf/number format/precision=1
    },
    yticklabel style={
        /pgf/number format/.cd,
        fixed,
        fixed zerofill,
        precision=3,
    },
    axis lines = left,
]

\addplot [
    color=mypurple, 
    line width=2pt, 
    mark=*, 
    mark options={fill=mypurple, scale=0.8, solid}
] coordinates {
    (0.2, 0.3408)
    (0.3, 0.3431)
    (0.4, 0.3444)
    (0.5, 0.3443)
    (0.6, 0.3438)
    (0.7, 0.3410)
};

\end{axis}
\end{tikzpicture}%
    \caption{}
    \label{fig:fix}
\end{subfigure}\hfill
\begin{subfigure}[]{0.32\linewidth}
    \centering
    \begin{tikzpicture}[scale=0.7]

\definecolor{myred}{RGB}{245, 70, 100}

\begin{axis}[
    width=6cm, 
    height=5cm,
    axis x line*=bottom,
    axis y line*=left,
    legend cell align={left},
    legend style={
        draw=none,
        at={(1.0, 0.05)},        
        anchor=south east,
        legend columns=1,
        fill=none,
        font=\small
    },
    xtick style={black},
    ytick style={black},
    xtick distance=10,     
    ytick distance=0.002,  
    ymin=0.301,            
    ymax=0.313,            
    tick align=outside,
    tick pos=left,
    xlabel={Sample Steps},
    y tick label style={
        /pgf/number format/.cd,
        fixed,
        fixed zerofill,
        precision=3,
        },
    axis lines = left,
]

\addplot [
    color=myred, 
    line width=2pt, 
    mark=*, 
    mark options={fill=myred, scale=0.8, solid}
] table {
    35 0.3030
    45 0.3044
    55 0.3057
    65 0.3066 
    75 0.3080
    85 0.3086
    95 0.3088
    105 0.3089
};
\addlegendentry{SDE}

\draw [dashed, line width=1.5pt, black!70] 
    (axis cs:35, 0.3114) -- (axis cs:105, 0.3114) 
    node[pos=0.5, above, font=\small, text=black] {ODE (sample steps = 35)};

\end{axis}
\end{tikzpicture}%
    \caption{}
    \label{fig:sde}
\end{subfigure}\hfill
\begin{subfigure}[]{0.32\linewidth}
    \centering
    \begin{tikzpicture}[scale=0.7]
\begin{axis}[
    width=5.8cm,
    height=4.7cm,
    ymin=0.3300, ymax=0.3460, 
    xtick={1, 2, 3, 4},
    xticklabels={Fixed-ratio, Adaptive, uSSP, uMulti}, 
    boxplot/draw direction=y,
    boxplot/box extend=0.4, 
    axis lines = left,
    enlarge x limits = true,
    enlarge y limits = true,
    xticklabel style={rotate=30},
    y tick label style={
        /pgf/number format/.cd,
        fixed,
        fixed zerofill,
        precision=3,
        },
]

\addplot [
    color=black,
    fill=blue!20, 
    mark=none,
    boxplot prepared={
        lower whisker=0.3332,
        lower quartile=0.3332,
        median=0.3410, 
        upper quartile=0.3444,
        upper whisker=0.3444,
    },
] coordinates {(0,0)};

\addplot [
    color=black,
    fill=red!20, 
    mark=none,
    boxplot prepared={
        lower whisker=0.3340,
        lower quartile=0.3340,
        median=0.3406,
        upper quartile=0.3448,
        upper whisker=0.3448,
    },
] coordinates {(0,0)};
 
\addplot [
    color=black,
    fill=green!20, 
    mark=none,
    boxplot prepared={
        lower whisker=0.3317,
        lower quartile=0.3317,
        median=0.3402,
        upper quartile=0.3433,
        upper whisker=0.3433,
    },
] coordinates {(0,0)};
 
\addplot [
    color=black,
    fill=orange!20, 
    mark=none,
    boxplot prepared={
        lower whisker=0.3360,
        lower quartile=0.3360,
        median=0.3388,
        upper quartile=0.3402,
        upper whisker=0.3402,
    },
] coordinates {(0,0)};

\end{axis}
\end{tikzpicture}%
    \caption{}
    \label{fig:p}
\end{subfigure}
\caption{
Further analyses.
(a)~The effect of different pruning retention ratios, where the number of initial candidates are scaled to keep the compute budget fixed.
(b)~SDE vs ODE performance for the Flux model.
Scaling up the number of SDE sampling steps from 35 to 105 fails to reach the ODE performance.
(c)~Box plot comparison of four pruning heuristics for the same compute cost: fixed-ratio (ours), adaptive-pruning (ours), unique SSP and unique multinomial.
}
\label{fig:analysis}
\end{figure}

\paragraph{Local or Global?}
First, we evaluate the effectiveness of SDE sampling on Flux (\cref{fig:sde}) on our validation set with reward models HPSv2.1.
We observe that the SDE consistently underperforms the ODE baseline, even when allocated triple the inference steps.
In the supplement, we further isolate the impact of compute allocation by comparing local search (refining existing candidates) against global search (exploring additional candidates).
We conclude that directing the compute budget toward broader global exploration yields better performance.

\paragraph{Cross-Model Generalisation.}
Since all diffusion and flow models progressively refine images from coarse to fine details, our Flux-finetuned reward model generalizes well to other architectures.
Empirical tests on SDXL \cite{sdxl} and SD3 \cite{sd3} confirm its consistent cross-model effectiveness, as shown in the supplement.

\paragraph{Limitations and Future Work.}
This work has several limitations.
First, TTS approaches have a reduced impact on models that are post-trained with human-preference rewards.
Second, our approach is less suitable for computationally-expensive reward models, such as VLM-based rewards \cite{nvila, unified_reward, vqascore, visionreward, hpsv3}.
Finally, our method requires that the generative latents be decoded into image space.
While the computational overhead of this is negligible for $\sim$30 steps flow models, it would be considerable for few-step shortcut models \cite{meanflow, song2023consistency, liu2023instaflow}.
A promising direction for future work is to develop a noise-aware reward model that operates directly in the latent space \cite{lpo, latentrewardtwostep, noiseclip}, thereby avoiding expensive VAE decoding and making our algorithm suitable for shortcut models.

\section{Conclusion}
\label{sec:conc}
We have demonstrated that, for ODE-based flow-matching models, (a)~global sampling with pruning is an effective test-time scaling strategy, (b)~sample diversity is critical for effective global search, and (c)~noise-aware rewards are critical for effective pruning.
Corresponding to these findings, we have proposed and evaluated a simple and effective global pruning method, a token repel mechanism for encouraging sample diversity, and an efficient self-distillation approach for finetuning reward models.
Our approach increases the search space under a fixed compute budget compared to existing methods, improving the expected quality of the final image.

\begin{ack}
\end{ack}

\bibliographystyle{abbrvnat}
\bibliography{main}

\newpage
\appendix

\section{Further Algorithmic Details}
The algorithm for our proposed global pruning approach is shown in \cref{alg:ttsp}.
The set of verification timesteps is denoted as $\mathcal{T} = \{\tau_1, \dots, \tau_{m-1}\} \subseteq [1, M-1]$, where $\tau_{m-1} = M-1$. $M$ is the final timestep.
The set of corresponding retention ratios is $\Gamma= \{\gamma_1, \dots, \gamma_{m-1}\}$, where $0<\gamma_i<1$ is the fraction of candidates that will be retained at timestep $\tau_i$.

Here, we show the budget analysis for the algorithm under the fixed-ratio pruning heuristic (retention ratios are predefined hyperparameters).
Let the computational cost of a single denoising step for one candidate be $B_d$, and that of a single verification be $B_v$.
For $N$ initial candidates, the total compute budget $B$ can be expressed as
\begin{align}
B &= \sum _{i=1}^{m} \prod_{j=0}^{i-1} \alpha_{j}  \left( (\tau_i - \tau_{i-1}) B_{d} + B_{v}\right) N,
\label{eq:cost}
\end{align}
where $\tau_0 := 0$ and $\alpha_0 := 1$.
Thus for a fixed budget $B$, the number of candidates that can explored is
\begin{align}
    N = \left\lfloor \frac{B}{\sum _{i=1}^{m} \prod_{j=0}^{i-1} \alpha_{j}
    \left( (\tau_i - \tau_{i-1}) B_{d} +B_{v}\right)} \right\rfloor.
    \label{eq:n}
\end{align}
Under the same compute budget $B$, our $N$ is much larger than that of the original best-of-$N$ strategy: $N = \lfloor B/(MB_d + B_v)\rfloor$ or beam searching strategy: $N = \lfloor B/(MB_d + |\mathcal{T}| B_v)\rfloor$.
The computational budget can be quantified in terms of total TFLOPs or GPU-hours. For instance, generating a $512 \times 512$ resolution image typically requires approximately 9.9 TFLOPs for the denoising process and 1.2 TFLOPs for the VAE decoding stage for Flux.

\begin{algorithm}[!t]
\caption{Proposed global search algorithm.}
\label{alg:ttsp}
\KwIn{
verification timesteps $\mathcal{T}$;
retention ratios $\Gamma$;
number of initial candidates $N$;
number of diffusion inference steps $M$;
diffusion denoising operation $\phi$;
VAE decoder $\nu$;
reward model $\psi$;
text prompt $p$
}
\KwOut{image $\mathcal{I}$}
$\mathcal{X} \gets \{x^i \mid  x^i \sim \mathcal{N}(0,I) , i \in [1, N]\} $ \\
\For{$j = 1$ \KwTo $M$}{
   $\mathcal{X} \gets \{x' \mid x' = \phi(x,p), x \in \mathcal{X} \} $ \\
   \uIf{$j = M$}{
    $\mathcal{R} \gets \{r \mid r = \psi(\nu(x),j), x \in \mathcal{X}\}$ \\
    $i \gets \argmax (\mathcal{R})$ \\
    $\mathcal{I} \gets \nu(\mathcal{X}_i)$ \\
   \Return $\mathcal{I}$
   }
  \uElseIf{$j \in \mathcal{T}$}{
    $\mathcal{R} \gets \{r \mid r = \psi( \nu(x),j), x \in \mathcal{X}\} $ \\
    $k \gets \Gamma_{\text{index}(j, \mathcal{T})} |\mathcal{R}|$ \\
    $\mathcal{X} \gets \{\mathcal{X}_i \mid i \in \argtopk(\mathcal{R}, k)\} $ \\
  }
 }
\end{algorithm}

\section{Scaling Compute}
Our approach enables effective test-time scaling.
By increasing the number of search candidates, generation performance can be consistently improved before reaching convergence.
\Cref{plot:scale} validates this trend across our validation set, showing that quality scales with more computational budget allocated.

\begin{figure}[!t]
    \centering
    \begin{subfigure}{0.32\textwidth}
        \centering
        \begin{tikzpicture}[scale=0.8]
\definecolor{myblue}{RGB}{70, 150, 220}
\begin{axis}[
    width=5cm, height=4cm,
    xlabel={Number of candidates ($N$)},
    xmin=0, xmax=22,
    xtick={3, 6, 9, 12, 15, 18, 21},
    ylabel={Reward},
    ymin=0.32, ymax=0.35,
    ytick={0.32, 0.33, 0.34, 0.35},
    axis x line*=bottom,
    axis y line*=left,
    legend style={draw=none, at={(0.97,0.05)}, anchor=south east, fill=none, font=\small},
    ymajorgrids=true,
    grid style={dashed,gray!30},
    tick align=outside, tick pos=left,
]
\addplot [color=myblue, line width=1.5pt, mark=*] table [row sep=newline]{
3	0.3234
6	0.3315
9	0.3360
12	0.3399
15	0.3430
18	0.3447
21	0.3466
};
\end{axis}
\end{tikzpicture}%
        \caption{HPSv2.1}
    \end{subfigure}\hfill
    \begin{subfigure}{0.32\textwidth}
        \centering
        \begin{tikzpicture}[scale=0.8]
\definecolor{myyellow}{RGB}{255, 180, 60}
\definecolor{myred}{RGB}{245, 70, 100}
\definecolor{myblue}{RGB}{70, 150, 220}
\definecolor{mygreen}{RGB}{60, 200, 110}
\begin{axis}[
    width=5cm, height=4cm,
    xlabel={Number of candidates ($N$)},
    xmin=0, xmax=22,
    xtick={3, 6, 9, 12, 15, 18, 21},
    ylabel={Reward},
    ymin=23.0, ymax=23.8,
    ytick={23.0, 23.2, 23.4, 23.6, 23.8},
    axis x line*=bottom,
    axis y line*=left,
    legend style={draw=none, at={(0.97,0.05)}, anchor=south east, fill=none, font=\small},
    ymajorgrids=true,
    grid style={dashed,gray!30},
    tick align=outside, tick pos=left,
]
\addplot [color=myyellow, line width=1.5pt, mark=*] table [row sep=newline]{
3	23.0705
6	23.3019
9	23.4526
12	23.5908
15	23.6616
18	23.6945
21	23.7341
};
\end{axis}
\end{tikzpicture}%
        \caption{PickScore}
    \end{subfigure}\hfill
    \begin{subfigure}{0.32\textwidth}
        \centering
        \begin{tikzpicture}[scale=0.8]

\definecolor{myyellow}{RGB}{255, 180, 60}
\definecolor{myred}{RGB}{245, 70, 100}
\definecolor{myblue}{RGB}{70, 150, 220}
\definecolor{mygreen}{RGB}{60, 200, 110}


\begin{axis}[
    width=5cm, 
    height=4cm,
    xlabel={Number of candidates ($N$)},
    xmin=0, xmax=22,
    xtick={3, 6, 9, 12, 15, 18, 21},
    ylabel={Reward},
    ymin=1.5, ymax=1.9,
    ytick={1.5, 1.6, 1.7, 1.8, 1.9},
    axis x line*=bottom,
    axis y line*=left,
    legend cell align={left},
    legend style={
        draw=none,
        at={(0.97,0.05)},
        anchor=south east,
        fill=none,
        font=\small
    },
    ymajorgrids=true,
    grid style={dashed,gray!30},
    xtick style={black},
    ytick style={black},
    tick align=outside,
    tick pos=left,
]

\addplot [
    color=mygreen, 
    line width=1.5pt, 
    mark=*, 
    mark options={fill=mygreen, solid}
] table [row sep=newline]{
3	1.561
6	1.6587
9	1.7355
12	1.7937
15	1.8494
18	1.8646
21	1.8698
};


\end{axis}
\end{tikzpicture}%
        \caption{ImageReward}
    \end{subfigure}%
    \caption{Scaling the compute with our algorithm by increasing the number of search candidates.
    The obtained image reward increases with the compute.
    }
    \label{plot:scale}
\end{figure}

\section{Additional Comparisons}

Another work, SDE-RBF \cite{tts_kim}, proposes a TTS pipeline designed for flow-matching generative models.
It first converts the SDE process with a variance-preserving scheduler as for classic diffusion models \cite{ddpm}, then leverages a greedy depth-first search process with a rollover-budget-forcing design.
We compared its performance with our method in \cref{tab:compare_rbf}, while maintaining the same total number of denoising steps.

As well as achieving better performance, our method offers two additional advantages.
\textbf{1. Efficiency}: SDE-RBF requires the reward verifier to run at each denoising step, introducing additional computation.
Also, in our method, denoising and verification processes can be run in parallel, where inference speeds are further enhanced by internal transformer optimizations.
Furthermore, as discussed in the main paper, SDE-based approaches generally require more inference timesteps to match the same image quality as ODE.
In practice, despite using the same denoising NFEs, their method requires $\sim 57$ seconds per prompt, while ours requires only $\sim 40$ seconds on the same hardware.
\textbf{2. Multi-reward flexibility}: Our method has better flexibility when it comes to multi-reward verification---called the ensemble reward in the main paper.
The rollover budget design in SDE-RBF requires evaluating each particle sequentially, whereas our method operates entirely in parallel.
This allows us to simply use the sum of ranks from multiple rewards as the verification output.
Conversely, their method requires a carefully designed heuristic function to merge ensemble rewards, as different reward models often possess varying scales and distributions.

\begin{table}[t]
    \centering
    \small
    \setlength{\tabcolsep}{4pt}
    \caption{
    Quantitative comparison with SDE-RBF with the same number of total denoising steps (approximately equivalent to fully denoising 6 images).
    }
    \label{tab:compare_rbf}
    \renewcommand{\arraystretch}{1.1}
    \begin{tabularx}{\linewidth}{@{} lc Y Y Y @{}}
        \toprule
        \multicolumn{2}{r}{Verifier Reward:} & HPS & PickScore & ImageReward  \\
        \multicolumn{2}{r}{Evaluation Reward:} & HPS & PickScore & ImageReward \\
        \midrule
        SDE-RBF \cite{tts_kim} & SDE & 0.3074 & 22.83 & 1.125 \\
        Ours & ODE  & \textbf{0.3314}  & \textbf{23.35} & \textbf{1.682} \\
        \bottomrule
    \end{tabularx}
\end{table}

\section{Cross-Model Generalisation for Noise-Aware Finetuning}
Since diffusion-based or flow-matching models share a common coarse-to-fine refinement process, reward models fine-tuned on the decoded intermediate images from one architecture can effectively generalize to others.
We validate our noise-aware Flux-finetuned reward models on SD3 \cite{sd3} and SDXL \cite{sdxl} with the same validation set.
As demonstrated in \cref{exp:cross_model}, the models yield visible gains in both Kendall coefficient and recall, confirming their robustness in cross-model scenarios.
That is, \textit{noise-aware reward fine-tuning is not strongly coupled to the generative model used to generate the training data}.

\begin{table}[!t]
    \centering
    \small
    \renewcommand{\arraystretch}{1.1}
    \setlength{\tabcolsep}{4pt}
    \caption{
    A noise-aware reward model, fine-tuned on Flux intermediate images (``w/ Flux NARF''), generalizes to other generative image models.
    We show that this finetuned reward model improves TTS performance for the SD3 and SDXL models, across different denoising timesteps $t$.
    }
    \label{exp:cross_model}
    \begin{tabularx}{\linewidth}{@{} lc CCC CCC @{}}
        \toprule
        & &  \multicolumn{3}{c}{Kendall coefficient $\uparrow$} & \multicolumn{3}{c}{Recall $\uparrow$} \\
        \cmidrule(lr){3-5} \cmidrule(l){6-8}
        Model & w/ Flux NARF & $t=21$ & $t=14$ & $t=7$ & $t=21$ & $t=14$ & $t=7$ \\
        \midrule
        Flux &  & 0.693 & 0.521 & 0.258  & 0.944  & 0.863 & 0.712 \\
        Flux & $\checkmark$ & \textbf{0.736} & \textbf{0.645} & \textbf{0.410} & \textbf{0.962} & \textbf{0.922} & \textbf{0.812} \\
        \midrule
        SD3-Medium &  & 0.708 & 0.535 & 0.253  & 0.927  & 0.833 & 0.637 \\
        SD3-Medium & $\checkmark$ & \textbf{0.721} & \textbf{0.576} & \textbf{0.351}  & \textbf{0.940}  & \textbf{0.861} & \textbf{0.739} \\
        \midrule
        SDXL &  & 0.591 & 0.463 & 0.274  & 0.888  & 0.799 & 0.679 \\
        SDXL & $\checkmark$  & \textbf{0.614} & \textbf{0.506} & \textbf{0.361}  & \textbf{0.909}  & \textbf{0.832} & \textbf{0.741} \\
        \bottomrule
    \end{tabularx}
\end{table}

\section{Comparing Local and Global Search}

In this section, we investigate how computation allocation between \textit{local search} (refining existing candidates) and \textit{global search} (exploring additional candidates) affects the performance of our algorithm.

Consider a simple two-stage pruning setup.
We generate $N$ candidates at the pruning timestep $d$, denoted as $X_d = \{x_d^1, \ldots, x_d^N\}$. Half of these candidates are pruned at this stage, and the final selection is performed at the end.
To study the effect of search strategies, we compare different ways of generating $X_d$, ranging from purely local search to purely global search.
Specifically, we initialize $X_d$ as an empty set.
We first denoise $m$ candidates to step $d$ (global exploration) and add them to $X_d$.
We then use the verifier to identify the best candidate in $X_d$, resample this latent $m$ times (local exploration), and add the resulting candidates to $X_d$. This process is repeated $k$ times such that $m \times k = N$.
When $m$ is large (and thus $k$ is small), the search strategy emphasizes local exploration; conversely, when $m$ is small (and $k$ is large), the strategy favors global exploration.
The total computation budget on denoising is constant across all settings.

Through this experiment, we aim to examine whether allocating more budget to global search or to local search leads to better performance under the same computational budget.
The results are demonstrated in \cref{plot:globallocal}.
First, we remove the proposed Repel mechanism and observe that varying the allocation between local and global search has little effect on the results, leading to a nearly flat performance curve.
We then introduce Repel in the global exploration stage (during the first iteration).
In this setting, we observe a clear difference: allocating more computation budget to global search yields better performance than spending the same budget on local refinement.

\begin{figure}[!t]
    \centering
    \begin{subfigure}{0.5\textwidth}
        \centering
        \begin{tikzpicture}[scale=0.9]

\definecolor{myblue}{RGB}{70, 150, 220}
\definecolor{myred}{RGB}{245, 70, 100}

\begin{axis}[
    width=8cm, 
    height=6cm,
    xlabel={Number of local search iterations ($K$)},
    xlabel style={font=\small},
    xmin=0.5, xmax=6.5,
    xtick={1, 2, 3, 4, 5, 6},
    xticklabels={1, 2, 3, 4, 6, 12}, 
    ylabel={Reward Score},
    ylabel style={font=\small},
    ymin=0.334, ymax=0.344,
    ytick={0.334, 0.336, 0.338, 0.340, 0.342, 0.344},
    yticklabel style={/pgf/number format/fixed, /pgf/number format/precision=3},
    axis x line*=bottom,
    axis y line*=left,
    legend cell align={left},
    legend style={
        draw=none,
        at={(0.97,0.95)},
        anchor=north east,
        fill=white,
        fill opacity=0.8,
        text opacity=1,
        font=\footnotesize
    },
    ymajorgrids=true,
    grid style={dashed,gray!30},
    tick align=outside,
    tick pos=left,
]

\addplot [
    color=myred, 
    line width=1.2pt, 
    mark=square*, 
    mark size=2pt,
    mark options={fill=myred, solid}
] table [row sep=newline]{
1	0.342156
2	0.340125
3	0.339790
4	0.338389
5	0.337434
6	0.337358
};
\addlegendentry{With Repel}

\addplot [
    color=myblue, 
    line width=1.2pt, 
    mark=*, 
    mark size=2pt,
    mark options={fill=myblue, solid}
] table [row sep=newline]{
1	0.3354
2	0.3362
3	0.3364
4	0.3358
5	0.3357
6	0.3365
};
\addlegendentry{Without Repel}

\node[anchor=south] at (axis cs: 1.2, 0.3425) {\tiny \textbf{$100\%$ Global}};
\node[anchor=south] at (axis cs: 5.8, 0.3378) {\tiny \textbf{$100\%$ Local}};

\end{axis}
\end{tikzpicture}%
    \end{subfigure}
    \caption{Investigating the effect of increasing the amount of local search performed under a fixed compute budget, from pure global search to pure local search, with and without the sample diversity-inducing Repel mechanism applied during global search.
    }
    \label{plot:globallocal}
\end{figure}

\section{Additional Qualitative Examples}

\begin{figure}[!t]
    \centering
    \begin{subfigure}{\linewidth}
        \centering
         \makebox[0.02\linewidth][c]{\raisebox{1em}{\rotatebox{90}{\small\textbf{Ours} \quad\quad\quad iSMC \quad\quad\quad T-Greedy \hspace{2pt} \quad\quad \  SoP \quad\quad\quad \ Greedy \quad \quad \quad BoN \quad\quad \ Random}}}
        \includegraphics[width=0.13\linewidth]{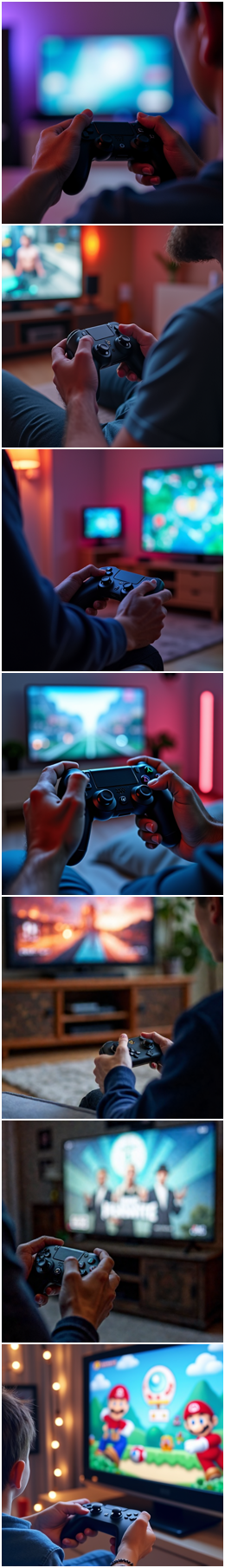}\hfill
        \includegraphics[width=0.13\linewidth]{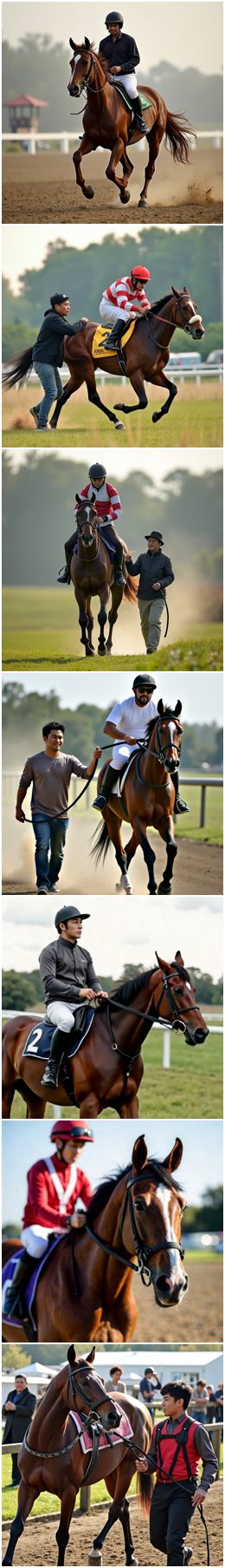}\hfill
        \includegraphics[width=0.13\linewidth]{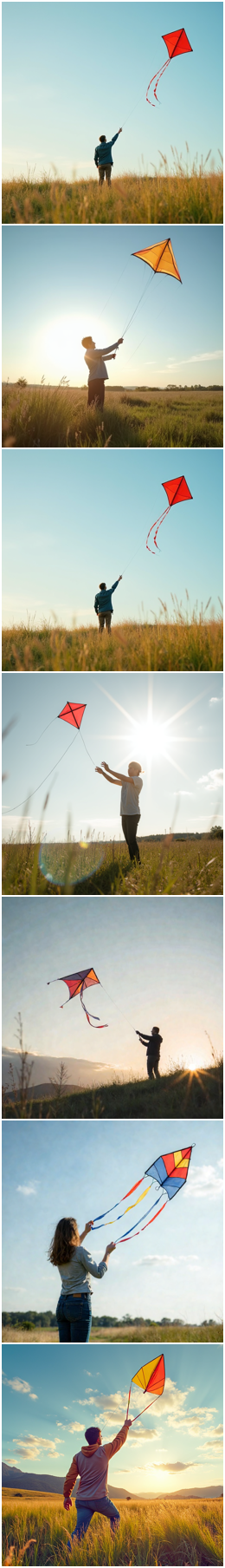}\hfill
        \includegraphics[width=0.13\linewidth]{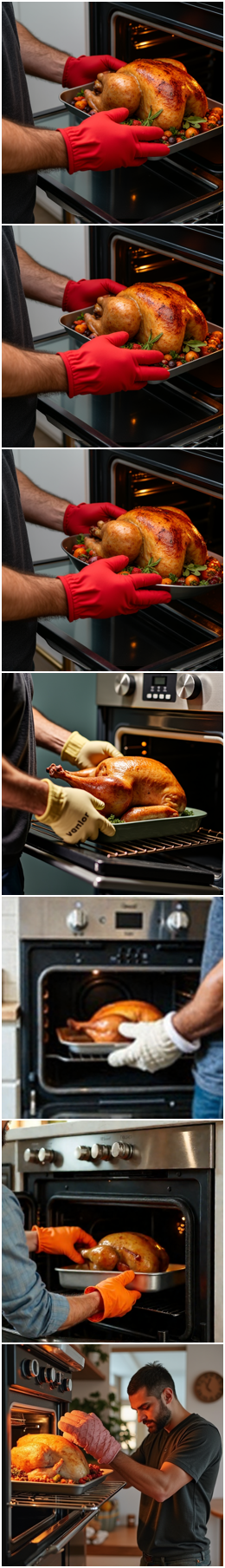}\hfill
        \includegraphics[width=0.13\linewidth]{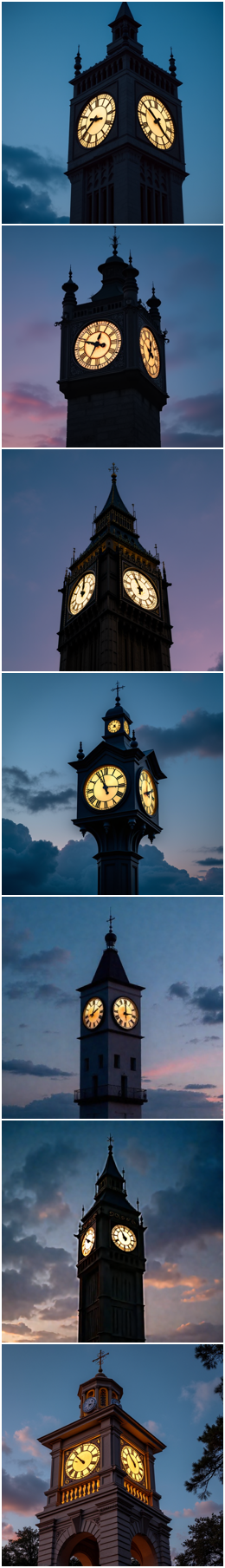}\hfill
        \includegraphics[width=0.13\linewidth]{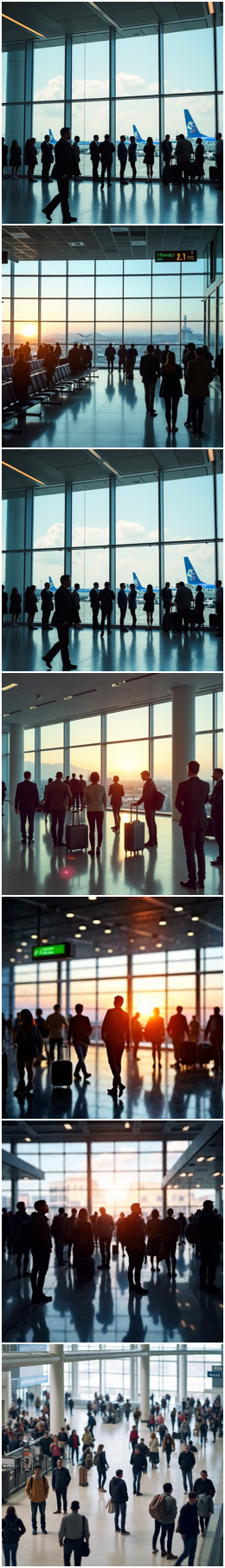}\hfill
        \includegraphics[width=0.13\linewidth]{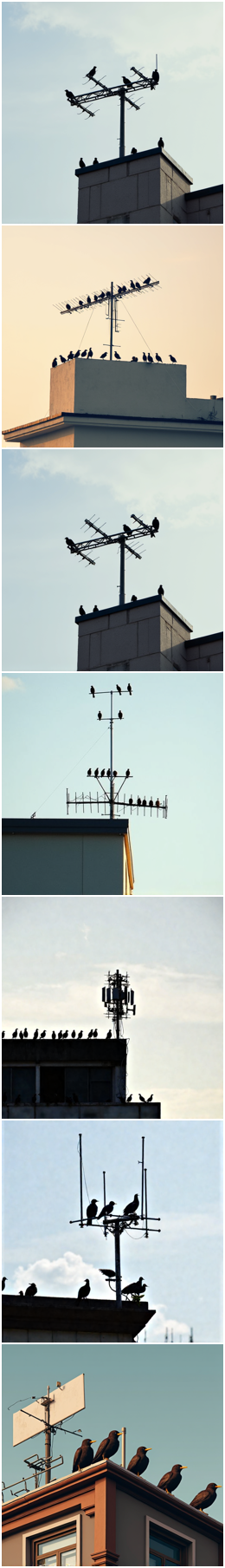}
    \end{subfigure}
    \caption{
        Qualitative results comparing TTS methods using the ensemble reward.
        Prompts from left to right:
        \textit{
        \hlc[pink!30]{1.} A person holding a remote while playing a game.
        \hlc[yellow!40]{2.} Racing horse being guided by an asian man.
        \hlc[green!20]{3.} A person flying a kite while standing in the grass.
        \hlc[cyan!20]{4.} A man placing a turkey in an oven with oven mitts.
        \hlc[pink!30]{5.} A clock tower with lighted clock faces, against a twilight sky.
        \hlc[yellow!40]{6.} A high shot of many people standing in an airport.
        \hlc[green!20]{7.} Group of birds sitting on top of a television antenna on a building.
        }
    }
    \label{fig:qualitative_supp}
\end{figure}

We provide additional qualitative results comparing our method with other approaches using the ensemble reward in \cref{fig:qualitative_supp}.

\section{Failure Cases}
\begin{figure}[!t]
    \centering
    \begin{subfigure}{\linewidth}
        \centering
        \makebox[0.02\linewidth][c]{\raisebox{1em}{\rotatebox{90}{\small\textbf{Ours} \quad\quad\quad Random}}}
        \includegraphics[width=0.23\linewidth]{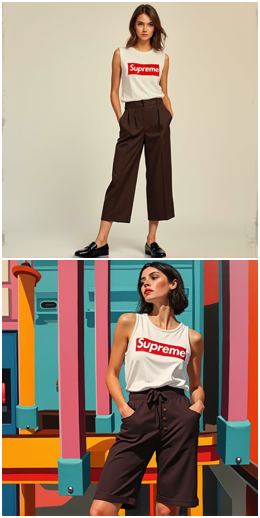}\hfill
        \includegraphics[width=0.23\linewidth]{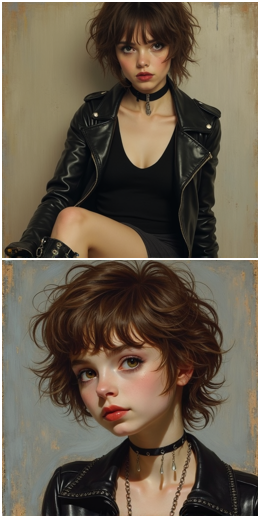}\hfill
        \includegraphics[width=0.23\linewidth]{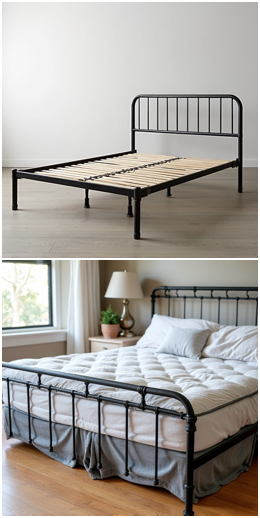}\hfill
        \includegraphics[width=0.23\linewidth]{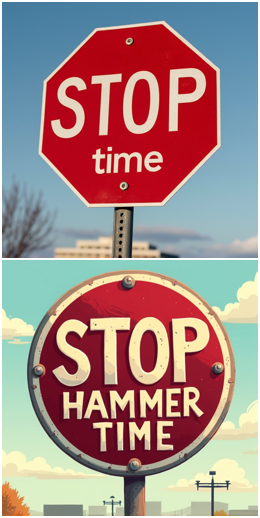}
    \end{subfigure}
    \caption{
        Some reward-hacking failure cases of our TTS method.
        Prompts from left to right:
        \textit{
        \hlc[pink!30]{1.} A painting of a woman by Zinaida Serebriakova wearing a T-shirt with the Supreme brand logo, a sleeveless white blouse, dark brown capris, and black loafers.
        \hlc[yellow!40]{2.} The image is a painting by Pierre-Auguste Renoir of an emo with short, messy brown hair, large entirely-black eyes, wearing a black tank top, leather jacket, knee-length skirt, choker, and boots.
        \hlc[green!20]{3.} A big metal bed frame with \textbf{no} mattress on it.
        \hlc[cyan!20]{4.} A stop sign with the phrase ``hammer time'' written on it.
        }
    }
    \label{fig:fail_supp}
\end{figure}

Our method's main bottleneck is its dependence on external reward models.
If the reward model has biases, our search process will expose them, a phenomenon known as reward hacking \cite{reward_hack}.
\Cref{fig:fail_supp} illustrates several failure cases where the model prefers complex backgrounds or ``big-headed'' figures.
Furthermore, the reward model struggles with complex text logic, often ignoring negations and incorrectly forcing labels onto objects.
Potentially, employing a more powerful reward model would mitigate these side effects.

\section{Related Work}
\label{sec:related}

\paragraph{Test-time Scaling.}
Test-time scaling aims to improve the performance of pre-trained models by allocating extra computation during inference~\cite{zhang2025surveytesttimescalinglarge}.
For large language models, a common strategy is to enhance the reasoning process \cite{OpenAIReasoning, guo2025deepseek}.
Chain-of-thought prompting~\cite{cot, cot2, cot3} generates intermediate reasoning steps to guide the final answer.
Subsequent works refined this by iteratively correcting outputs~\cite{madaan2023self, simples1} or decomposing complex questions into simpler, sequential sub-problems~\cite{zhou2023leasttomost}.
Another approach samples multiple candidate solutions in parallel and selects the optimal one (\ie, best-of-$N$)~\cite{ttscode, moneyllmtts, snellscale}.
Test-time scaling also extends to vision--language models.
EfficientTTS~\cite{Kaya2025EfficientTTS} performs inference-time adaptation by jointly optimizing image and text prompts.
VisVM~\cite{wang2025scaling} samples multiple reasoning paths and selects the best via a reward model.
Recently, test-time scaling has been applied to diffusion models \cite{tts_ma, tts_var, flowmap, fk, classic, tts_kim, tts_tut, tts_vidoet1}, including searching over initial noise~\cite{tts_ma, sana1_5}, exploring noise trajectories~\cite{tts_greedy, dynamic_search}, maximizing probability density along the denoising path~\cite{fk, tts_tut}, and iteratively refining the conditional input~\cite{reflect_perfect, reflectdit, tts_kim_p}.
Global search over the noise prior has the widest search space, but incurs a high computational cost.
In contrast, we focus on improving TTS efficiency in this setting through early-stage pruning.

\paragraph{Preference Alignment in Diffusion Models.}
The goal of preference alignment is to guide diffusion models toward outputs that better satisfy task objectives and align with human preferences.
Post-training approaches typically leverage reward models via reinforcement learning~\cite{ppo, ddpo, sdrl, alignt2ihf, clarkdirectly, imagereward} or directly optimize diffusion models with human feedback~\cite{dpo, d3po, wu2024multimodal}.
For example, GRPO-based methods~\cite{xue2025dancegrpo, flowgrpo} employ efficient gradient regularization to balance GPU memory usage and training stability while improving alignment quality.
DPO-style methods further enhance preference alignment through step-aware preference modeling~\cite{spo, lpo}.
Alternatively, some techniques achieve preference alignment during inference, without modifying the diffusion models.
Some methods perform test-time optimization by refining the initial noise distribution~\cite{golden, hypernoise, noise_reno, noise_worth}, while others intervene directly in the denoising trajectory~\cite{dymo}.
Meanwhile, the inherent stochasticity of diffusion models enables search-based strategies.
In this paper, we investigate how to align rewards for flow matching models at test time within this search paradigm.

\end{document}